\documentclass[sigconf]{acmart}

\usepackage{bm}
\usepackage{graphicx}
\usepackage{float}
\usepackage[utf8]{inputenc}
\usepackage{enumitem}
\usepackage{amsfonts}
\usepackage[ruled,linesnumbered, noend]{algorithm2e}
\usepackage{pgfplots}
\usepackage{makecell}
\usepackage{arydshln}
\usepackage{subcaption}
\usepackage{tikz}
\usepackage{adjustbox}
\usepackage{booktabs}
\usepackage{multirow}
\usepackage{threeparttable}
\usepackage{comment}
\usepackage{textcomp}
\usepackage{mathtools}
\usepackage{algorithmic}
\usepackage{subcaption}
\usepackage[marginal]{footmisc}

\newcommand\M{GOAL}
\newcommand\subM{SCPO}

\usepackage{newfloat}
\usepackage{listings}

\AtBeginDocument{%
  }

\copyrightyear{2026}
\acmYear{2026}
\setcopyright{cc}
\setcctype{by}
\acmConference[KDD '26]{Proceedings of the 32nd ACM SIGKDD Conference on Knowledge Discovery and Data Mining V.2}{August 09--13, 2026}{Jeju Island, Republic of Korea}
\acmBooktitle{Proceedings of the 32nd ACM SIGKDD Conference on Knowledge Discovery and Data Mining V.2 (KDD '26), August 09--13, 2026, Jeju Island, Republic of Korea}
\acmDOI{10.1145/3770855.3818423}
\acmISBN{979-8-4007-2259-2/2026/08}

\begin{document}
\setcounter{secnumdepth}{3}

\title{Generative Optimization for Incentivized Advertising with  Global Level Constraints}

\author{Gege Chen}
\authornote{Contributed equally to this research.}
\authornote{Work done during an internship at Kuaishou Technology.}
\affiliation{
  \institution{University of Electronic Science and Technology of China}
  \city{Chengdu}
  \country{China}
}
\email{ggchen@std.uestc.edu.cn}

\author{
    Ning Luo
    }
\authornotemark[1]
\affiliation{
  \institution{Kuaishou Technology} 
  \city{Beijing} 
  \country{China}
}
\email{luoning@kuaishou.com}

\author{Hao Jiang}
\authornotemark[1]
\affiliation{
  \institution{Kuaishou Technology}
  \city{Beijing} 
  \country{China}
}
\email{jianghao10@kuaishou.com}

% \author{Da Li \and
%         Wenzheng Shu
% }
% \affiliation{
%   \institution{Kuaishou Technology}
%   \city{}
%   \country{}
% }
% \email{{lida06,shuwenzheng}@kuaishou.com}

\author{Da Li}

\affiliation{
  \institution{Kuaishou Technology}
  \city{Beijing}
  \country{China}
}
\email{lida06@kuaishou.com}

\author{Wenzheng Shu}

\affiliation{
  \institution{Kuaishou Technology}
  \city{Beijing}
  \country{China}
}
\email{shuwenzheng@kuaishou.com}

\author{Teng Sha}

\affiliation{
  \institution{Kuaishou Technology}
  \city{Beijing}
  \country{China}
}
\email{shateng@kuaishou.com}

\author{Yanxiang Zeng}
\affiliation{
  \institution{Kuaishou Technology}
  \city{Beijing}
  \country{China}
}
\email{zengyanxiang@kuaishou.com}

\author{
        Wenxin Tai
}
% \authornote{Corresponding Author.}
\affiliation{
  \institution{University of Electronic Science and Technology of China}
  \city{Chengdu}
  \country{China}
}
\email{amperetai@gmail.com}

\author{
        Fan Zhou
}
% \authornotemark[2]
\affiliation{
  \institution{University of Electronic Science and Technology of China}
  \city{Chengdu}
  \country{China}
}
\email{fan.zhou@uestc.edu.cn}        

\author{Xialong Liu}
% \authornote{Corresponding author.}
\affiliation{
  \institution{Kuaishou Technology}
  \city{Beijing}
  \country{China}
}
\email{zhaolei16@kuaishou.com}

\renewcommand{\shortauthors}{Gege Chen et al.}

\begin{abstract}

Incentivized advertising allocates monetary or virtual rewards to drive user engagement, where a key challenge is optimizing continuous incentive magnitudes under strict global constraints. This problem is complicated by high-frequency interactions, delayed feedback, and non-Markovian user dynamics such as fatigue, which limit the effectiveness of existing uplift modeling and constrained reinforcement learning approaches. To address these challenges, we propose GOAL, a constraint-aware generative framework that formulates incentive allocation as a conditional sequence generation problem. GOAL directly generates incentive magnitudes conditioned on user histories and system-level global pressure, and integrates a hierarchical causal state encoder to capture both local behavioral dynamics and long-range dependencies. To enable flexible constraint control, we introduce \textbf{S}afe \textbf{C}onstrained \textbf{P}olicy \textbf{O}ptimization (SCPO), which learns a single generative policy that generalizes across a spectrum of ROI constraints without retraining.
Experiments on large-scale real-world data and a synthetic fatigue-aware environment show that GOAL improves long-term revenue and user retention while substantially reducing ROI violation rates compared to strong baselines.

\end{abstract}

\authornote{Corresponding author.}

\maketitle

\section{Introduction}

\begin{figure}[t]
  \centering
  \includegraphics[width=0.9\columnwidth]{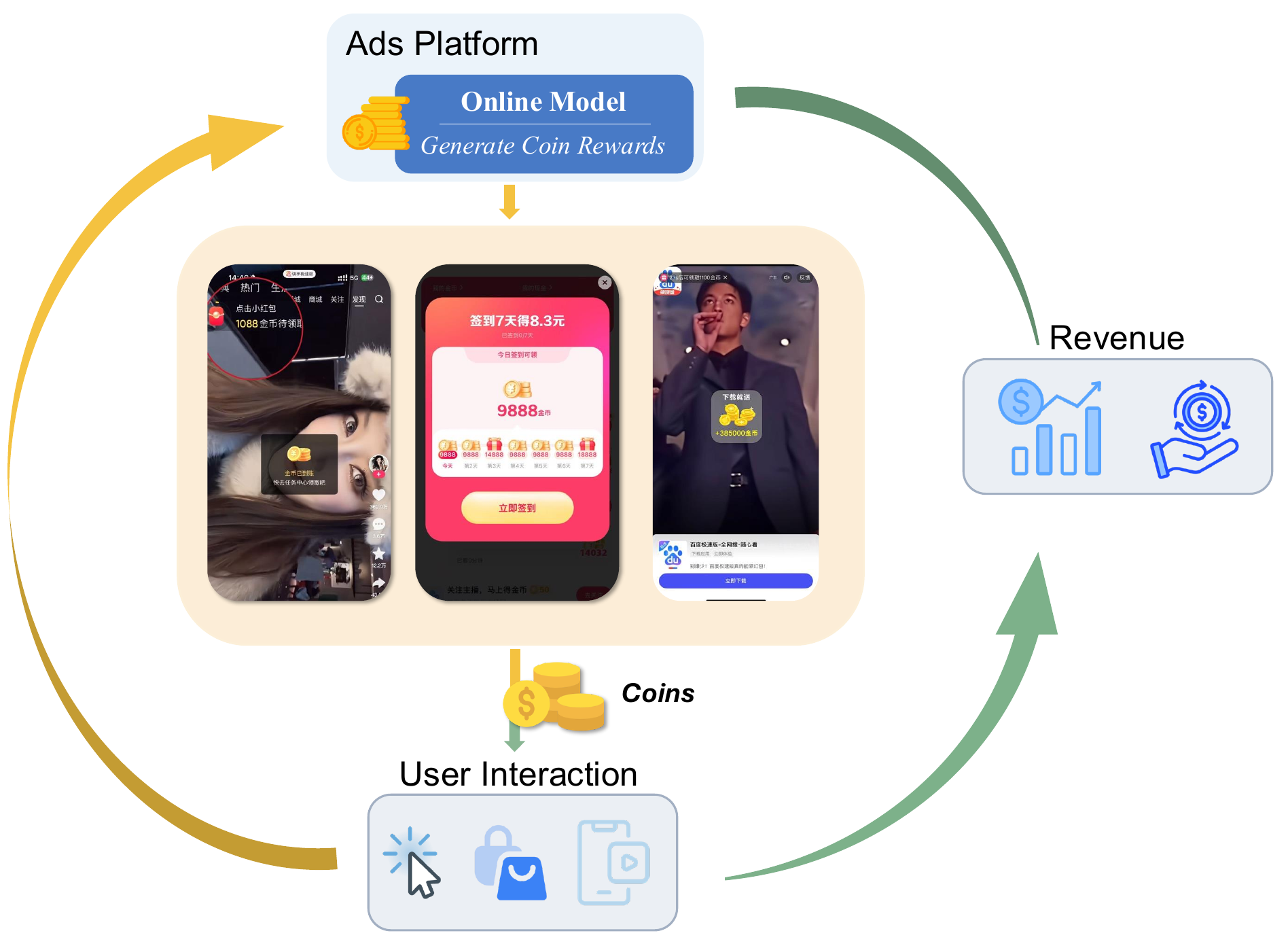} 
  \caption{
  Overview of the incentivized advertising framework. The platform leverages online models to directly generate incentive amounts based on user context and historical behaviors, distributing them to users to encourage participation in ad interactions. Corresponding user interaction histories are collected to enable the platform's continuous optimization under cost and performance constraints.}
  \vspace{-4mm}
\label{fig:intro}
\end{figure}

Incentivized advertising has become a core mechanism for driving user engagement in modern recommender and advertising systems, where platforms dynamically allocate monetary incentives (e.g., virtual coins or coupons) to stimulate user interactions~\cite{zhang2021bcorle, chen2022bcrlsp}. Unlike conventional ad recommendation settings that focus solely on content selection~\cite{deng2025onerec, rajput2023recommender}, incentivized advertising requires jointly deciding how much incentive to allocate for each interaction, under strict global constraints~\cite{wu2018budget, goldenberg2020free} as Fig.~\ref{fig:intro} shows. Excessive incentives may yield short-term engagement gains but severely harm long-term profitability, while overly conservative strategies fail to sustain user participation. Designing an incentive allocation policy that balances long-term value and financial safety remains a fundamental challenge in large-scale industrial systems.

% Modern advertising systems operate under increasingly complex constraints, where platforms must simultaneously optimize user engagement, advertiser value, and long-term ecosystem sustainability~\cite{lin2019pareto}. Incentivized Advertising has emerged as a widely adopted paradigm to enhance user engagement and retention, in which platforms decide the magnitude of monetary incentives (e.g., virtual coins) for each user interaction, conditioned on user context and historical behavior. Unlike conventional advertising settings, incentivized advertising requires not only selecting appropriate ad content but also determining the incentive intensity tailored to individual user contexts. Critically, this decision process is subject to strict Return on Investment (ROI) constraints~\cite{wu2018budget, goldenberg2020free}: excessive incentives may boost short-term engagement but can severely undermine long-term profitability, while overly conservative strategies fail to sustain user participation.

From a decision-making perspective, incentive allocation can be viewed as a sequential optimization problem with delayed and cumulative effects. Each incentive decision not only affects immediate user response but also influences future behavior through latent dynamics such as user fatigue and incentive saturation. However, existing approaches struggle to address this challenge effectively. Uplift modeling methods, widely adopted in production systems, focus on short-term treatment effects~\cite{pei2019value, zou2019reinforcement, zheng2018drn} and rely on two-stage pipelines that are highly sensitive to prediction errors, leading to sub-optimal decisions under constraints such as Return on Investment (ROI)~\cite{hansotia2002direct, zhao2019unified, goldenberg2020free}. Offline reinforcement learning (RL) methods explicitly optimize long-term objectives, but they typically rely on Markovian assumptions that are violated in high-frequency user interactions~\cite{hausknecht2015deep, zhao2018deep, li2024modeling} and tend to learn overly conservative policies due to distributional shift and limited exploration~\cite{kiyohara2021accelerating, korenkevych2024offline, liu2025session}.

Recently, generative sequence modeling paradigms inspired by large language models have shown promise in modeling complex user behaviors by directly generating actions conditioned on historical trajectories~\cite{geng2022recommendation, deng2025onerec, guo2025onesug, tay2022transformer}. Despite their expressive power, standard Transformer-based architectures exhibit a critical limitation in incentivized advertising~\cite{li2023survey, xiao2021early}: decision-critical signals, such as abrupt user fatigue or intent shifts, are often highly localized in time, while global self-attention tends to dilute these signals~\cite{zhou2021informer, liu2024lost} when processing long interaction histories. Moreover, existing generative alignment methods primarily optimize unconstrained scalar rewards, making them ill-suited for strict, system-level constraints that couple decisions across time~\cite{xiao2021early}.

To address the aforementioned challenges, we introduce \textbf{\M}, a generative optimization framework with global-level constraints, specifically tailored for incentivized advertising. Our improvements primarily focus on enhancing generative decision models along two critical dimensions: (i)\textbf{capturing non-Markovian and localized user dynamics in long interaction histories}, and (ii) \textbf{enforcing strict global-level ROI constraints within a unified generative decision framework.} Concretely, \M\ integrates a hierarchical causal state encoder that explicitly models dense local behavioral dynamics before performing global temporal reasoning, together with a constraint-conditioned generative policy that adapts incentive strategies according to system-level pressure. On top of this architecture, we introduce Safe Constrained Policy Optimization (\subM), which trains a single generative policy over a distribution of Lagrange multipliers. By separating and normalizing preference signals under each constraint level during alignment, \subM\ enables the learned policy to adapt its incentive strategy to different global constraint targets at inference time without retraining. To summarize, our main contributions are as follows:

\begin{itemize}[leftmargin=*]
    \item \textbf{Generative Framework for Incentivized Advertising.} We propose \M\ and explore the application of a generative paradigm to incentivized advertising, enabling the model to capture finer-grained behavioral patterns as well as both short and long-term dependencies.

    \item \textbf{\textit{\subM} Alignment Strategy.}
    We propose \textit{\subM}, a constraint-aware alignment algorithm for safe exploration. Unlike GRPO, which may induce \textit{revenue-cost imbalance} by solely maximizing system value, \textit{\subM} integrates generative policy optimization with Lagrangian multiplier to strictly enforce global constraints while optimizing long-term ecosystem value.

    \item \textbf{Validation on Industrial Scale.} We validate the superior performance of our approach on a diverse set of real-world industrial and synthetic datasets. The results demonstrate significant improvements in core metrics.
\end{itemize}

\section{Preliminary}
In this section, we formally formulate the incentive decision problem as a generative sequence modeling task and introduce the constrained optimization framework based on Lagrangian duality.
\subsection{Problem Formulation}
\label{subsec:task}

Given a dataset $\mathcal{D} = \{(\mathcal{H}_i, \lambda_i, a_i)\}_{i=1}^N$, where $N$ is the dataset size. For the $i$-th example (corresponding to a specific decision moment $t$ for a user), $\mathcal{H}_i$ represents the historical interaction trajectory derived via a sliding window of length $L$: $\mathcal{H}i = [\mathbf{x}_{t-L}, \dots, \mathbf{x}_{t-1}]$,where each composite event $\mathbf{x}_\tau = (\mathbf{s}_\tau, a_\tau) \in \mathcal{S} \times \mathcal{A}$ consists of the high-dimensional state features $\mathbf{s}_\tau$ and the received incentive $a_\tau$ at step $\tau$. Additionally, $\lambda_i \in \mathbb{R}^+$ represents the system-level constraint multiplier reflecting the real-time system pressure, and $a_i \in \mathcal{A}$ is the ground-truth incentive allocated at the current step $t$.

To harness the sequential reasoning capabilities of generative architectures for continuous control, we reformulate the incentive decision as an autoregressive generation task. Specifically, we introduce a discrete vocabulary $\mathcal{V}$, where each token represents a quantized numerical unit. We decompose the continuous target action $a_i$ into a sequence of tokens $\boldsymbol{g}_i = (g_i^1, g_i^2, \dots, g_i^{T_i})$, where $g_i^k \in \mathcal{V}$ denotes the token at the $k$-th generation step, and $T_i$ denotes the sequence length. Conversely, we design a deterministic mapping function $\phi(\cdot)$ that reconstructs the original scalar value $a_i$ from $g_i$, i.e., $a_i = \phi(g_i)=\sum_{k=1}^{T_i} \phi(g_i^k) \in \mathbb{R}^+$, where $\phi(g_i^k)$ denotes the numerical magnitude of the token. The details of vocabulary construction and mapping are presented in Section~\ref{tokenizer}. Our goal is to train a generative policy $\pi_\theta$ which, given the user history and constraint characteristics $(\mathcal{H}_i, \lambda_i)$, generates the corresponding predicted token sequence $\boldsymbol{\hat{g}}_i = (\hat{g}_i^1, \hat{g}_i^2, \dots, \hat{g}_i^{T_i})$. In turn, the predicted incentive value $\hat{a}_i$, reconstructed via $\hat{a}_i=\phi(\hat{g}_i)=\sum_{k=1}^{T_i} \phi(\hat{g}_i^k)$, approximates the actual incentive $a_i$ while adhering to the cost-efficiency trade-off governed by $\lambda_i$.

\begin{figure*}[t]
  \centering
  \includegraphics[width=\textwidth]{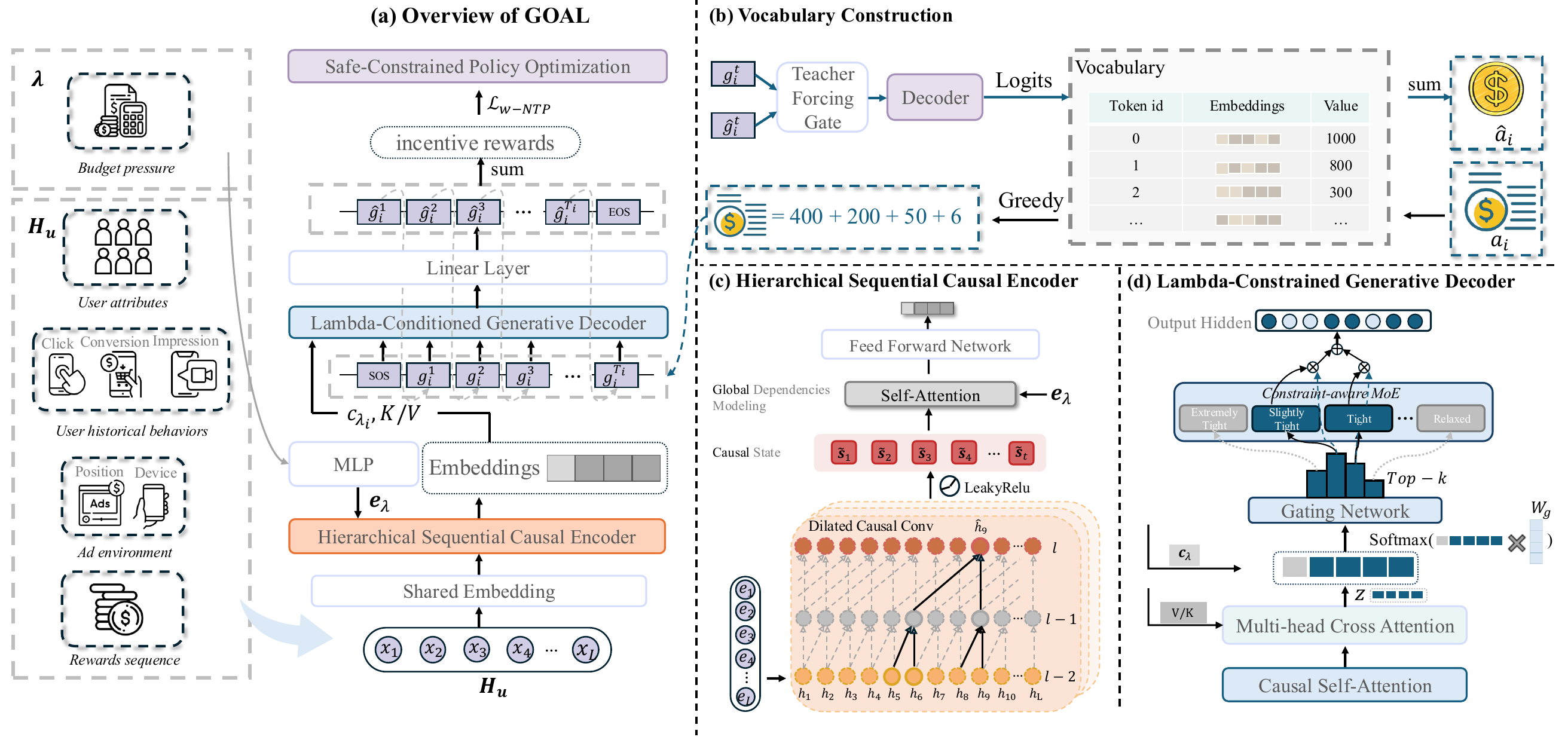}
  \caption{The overall architecture of~\M. (1) Hierarchical Causal Encoder: synergizes DCC and self-attention to capture both high-frequency local dynamics and global temporal dependencies. (2) $\lambda$-Conditioned Generative Decoder: replaces standard FFNs with MoE to enable the experts to perceive different levels of $\lambda$-induced constraint pressure. (3) Optimization: optimized via a weighted NTP objective $\mathcal{L}_{W-N T P}$.}
  \label{fig:framework}
  % \vspace{-4mm}
\end{figure*}

\subsection{Primal-Dual Constrained Policy Optimization}
\label{subsec:dual}
While the generative policy maximizes user value given the system-level state $\lambda$, the definition of $\lambda$ is governed by global business requirements, particularly strict ROI constraints in incentive scenarios.
\\
\textbf{Primal Problem.}
Consider a trajectory of \(N\) interactions, where action \(a_i\) incurs cost \(c_i\) and yields revenue \(r_i\). 
The global ROI is defined as $\frac{\sum_{i=1}^N r_i}{\sum_{i=1}^N c_i}$, which is required to exceed a minimum threshold \(\tau\).
The resulting reward decision problem is formulated as
\begin{equation}
\label{eq:primal_problem}
\max_{\{a_i\}} \sum_{i=1}^N r_i \quad \text{s.t.} \quad \frac{\sum_{i=1}^N r_i}{\sum_{i=1}^N c_i} \ge \tau, \quad 0 \le c_i \le C_{\max}.
\end{equation}
The fractional ROI constraint couples decisions across time, making direct optimization challenging.
\\
\textbf{Lagrangian Duality and Decoupling.}
We leverage the Lagrangian duality framework to decouple the global constraint. Specifically, we linearize the ROI constraint and introduce a Lagrange multiplier $\lambda^\prime \ge 0$ to incorporate it into the optimization objective as:
\begin{equation}
\begin{split}
    \mathcal{L}(\pi, \lambda^\prime) &= \sum_{i=1}^N r_i + \lambda^\prime \left( \sum_{i=1}^N r_i - \tau \sum_{i=1}^N c_i \right) \\
    &= (1 + \lambda^\prime) \sum_{i=1}^N r_i - \lambda^\prime\tau \sum_{i=1}^N c_i.
\end{split}
\end{equation}
Optimizing this objective is mathematically equivalent to maximizing the normalized return: \begin{equation} \max_{\pi \in \Pi} \mathcal{L}(\pi, \lambda) = \max_{\pi \in \Pi} \sum_{i=1}^N \left( r_i - \lambda c_i \right), \quad \text{where } \lambda = \frac{\lambda^\prime \tau}{1 + \lambda^\prime}. 
\label{eq:dual}
\end{equation}
Under the mild assumption that a feasible policy exists, strong duality holds. Thus, the primal problem is equivalent to the Lagrangian dual: $\min_{\lambda \in \mathbb{R}^+} \max_{\pi \in \Pi} \mathcal{L}(\pi, \lambda)$.

Therefore, given the optimal dual variable $\lambda^*$, the optimization objective reduces to:
\begin{equation}
    \max_{\pi} \sum_{i=1}^N (r_i - \lambda^* c_i).
\label{eq:final_op}
\end{equation}

\section{Methodology}
\label{sec:methodology}
As illustrated in Figure~\ref{fig:framework}, we propose GOAL, departing from conventional two-stage incentive optimization pipelines by unifying generation, constraint modeling, and policy optimization into a single generative framework. 
The model operates in two tightly coupled stages. First, a causal state encoder captures the temporally consistent evolution of user interests, while a Lagrangian multiplier embedding is injected as a global control signal to explicitly parameterize constraint tightness. Second, an autoregressive decoder equipped with constraint-aware MoE routing generates incentive sequences, dynamically switching between conservative and aggressive strategies based on the constraint context. On top of this architecture, we introduce a $\lambda$-generalized safe constrained policy optimization scheme (SCPO), enabling a single trained model to adapt to diverse ROI constraints. Together, these components embed constraint awareness into both the representation space and the optimization dynamics, yielding a flexible and scalable solution to constrained incentive generation.
In the following subsections, we elaborate on the implementation details of these components.

\subsection{Incentive Vocabulary Construction}
\label{tokenizer}
% Incentive allocation differs from language generation in that it involves continuous numerical values with strict magnitude constraints. Existing tokenization~\cite{wallace2019nlp} or bucketing methods~\cite{wang2021tuta, jie2022learning} fail to preserve numerical precision, especially for long-tailed incentive distributions. We therefore adopt a data-driven vocabulary construction approach based on dynamic quantile adjustment~\cite{ma2024generative}, the detailed process is outlined in Algorithm~\ref{alg:vocab_construction}.

\begin{algorithm}[t]
\caption{Incentive Vocabulary Construction}
\label{alg:vocab_construction}
\begin{algorithmic}[1]
\REQUIRE Incentive dataset $\mathcal{A} = \{a_j\}_{j=1}^N$, start percentile $q_{\text{start}}$, end percentile $q_{\text{end}}$, decay rate $\delta$, restoration error threshold $\epsilon_1$, and minimal precision threshold $\epsilon_2$.
\ENSURE Constructed vocabulary $\mathcal{V}$
\STATE Initialize an empty vocabulary $\mathcal{V} \leftarrow \{\}$
\STATE Sort $\mathcal{A}$ in descending order to obtain $\hat{\mathcal{A}} = \{\hat{a}_j\}_{j=1}^N$
\STATE Initialize iteration counter $i \leftarrow 1$, error metric $err \leftarrow \infty$, and current percentile $q \leftarrow q_{\text{start}}$
\WHILE{$err > \epsilon_1$}
    \STATE Compute the $q$-percentile value $o_i$ of the residual set $\hat{\mathcal{A}}$
    \IF{$o_i \le \epsilon_2$}
        \STATE \textbf{break} \COMMENT{Terminate if the token precision falls below the minimum threshold}
    \ENDIF
    \STATE Generate a new token $w_i$ such that $\phi(w_i) = o_i$
    \STATE $\mathcal{V} \leftarrow \mathcal{V} \cup \{w_i\}$ \COMMENT{Insert the unique token into the vocabulary}
    \STATE Update the residual values in $\hat{\mathcal{A}}$:
    \FOR{$j = 1$ \TO $N$}
        \IF{$\hat{a}_j \ge o_i$}
            \STATE $\hat{a}_j \leftarrow \hat{a}_j - o_i$
        \ENDIF
    \ENDFOR
    \STATE Update the error metric: $err \leftarrow \max_{1 \le j \le N} \left\{ \frac{\hat{a}_j}{a_j} \right\}$
    \STATE Update the percentile with decay rate $\delta$: $q \leftarrow \max(q \cdot \delta, q_{\text{end}})$
    \STATE $i \leftarrow i + 1$
\ENDWHILE
\RETURN $\mathcal{V}$
\end{algorithmic}
\end{algorithm}

Incentive allocation differs from language generation in that it involves continuous numerical values with strict magnitude constraints, where existing tokenization~\cite{wallace2019nlp} or bucketing methods~\cite{wang2021tuta, jie2022learning} fail to preserve numerical precision. This limitation stems from two inherent characteristics of the action space: a wide value range (e.g., 1 to 3,000 coins) and a heavy-tailed distribution. Standard one-step regression faces a severe optimization bottleneck under such a vast range, as uniform losses struggle to balance low-value regions with high-magnitude scales. Moreover, due to the long-tailed nature of marketing traffic, regression models are easily overwhelmed by high-frequency low-value samples, leading to severe underfitting on rare but decision-critical large incentives. To address these challenges, we adopt a data-driven vocabulary construction approach based on dynamic quantile adjustment~\cite{ma2024generative}, as detailed in Algorithm~\ref{alg:vocab_construction}.

The constructed vocabulary $\mathcal{V} = \{w_1, w_2, \dots, w_{|\mathcal{V}|}\}$ guarantees that each token is unique, enabling the comprehensive representation of the continuous incentive space using a finite set of discrete units. To transform a specific incentive $a_i$ into a token sequence $\boldsymbol{g}_i = (g_i^1, g_i^2, \dots, g_i^{T_i})$, we must ensure that the original value is reconstructible with bounded error while minimizing the sequence length $M$. To achieve these dual objectives, we employ a deterministic greedy decomposition algorithm that iteratively decomposes the total reward $a_t$ by always selecting the largest available token from $\mathcal{V}$ less than or equal to the current residual value. Crucially, this strategy naturally enforces a monotonicity constraint $\phi(g_i^1) \ge \phi(g_i^2) \ge \dots \ge \phi(g_i^{T_i})$. This captures a hierarchical coarse-to-fine structure, ensuring that the leading tokens encode the dominant magnitude of the incentive, while subsequent tokens efficiently handle residual precision.

\subsection{Autoregressive Incentive Generation}
\label{subsec:generation}
The generation module serves as the actor in our framework, which maps the composite state inputs into a precise incentive action.
\subsubsection{Hierarchical Causal Encoder}
\label{subsubsec:state_encoding}

Distinct from conventional encoders that directly feed raw embeddings into Transformer layers, we introduce a specialized causal state encoder module to explicitly capture the dynamic causality between historical behaviors and current states. The encoding process operates sequentially, moving from initial feature projection to causal modeling, and finally to global context integration.

Given the user historical trajectory $\mathcal{H}_u = [\boldsymbol{x}_{t-L}, \dots, \boldsymbol{x}_{t-1}]$ defined in the problem formulation, which corresponds to the sequence $\left(\boldsymbol{x}_{1}, \ldots, \boldsymbol{x}_{L}\right)$ illustrated in Fig.~\ref{fig:framework}, we first project these features into dense continuous vectors. This process yields an initial embedding sequence $\boldsymbol{E} = (\boldsymbol{e}_1, \dots, \boldsymbol{e}_L)$, which serves as the input to the causal module. To simulate the user's dynamic interests while strictly adhering to temporal causality, we employ a Dilated Causal Convolution (DCC) block~\cite{bai2018empirical}. This module aggregates previous behaviors to form a causal state representation. Let $\boldsymbol{H}^{(0)} = \boldsymbol{E}$ be the input to the DCC. We generate the hidden representations $\boldsymbol{H}^{(l)}=\left(\boldsymbol{h}_1^{(l)}, \cdots, \boldsymbol{h}_L^{(l)}\right)$ for the $l$-th convolutional layer via DCC:
\begin{equation}
    \boldsymbol{H}^{(l)} = \text{DilatedCausalConv}(\boldsymbol{H}^{(l-1)}), \quad l = 1, \dots, L_{1}.
\end{equation}
Specifically, the hidden state at time $t$, denoted as $\boldsymbol{h}^{(l)}_t$, is computed by convolving filter weights $f$ with past states:
\begin{equation}
    \boldsymbol{h}^{(l)}_t = \sum_{i=0}^{k_c - 1} f(i) \cdot \boldsymbol{h}^{(l-1)}_{t - d_c \cdot i},
\end{equation}
where $d_c$ and $k_c$ represent the dilation factor and filter size, respectively. This formulation ensures that $\boldsymbol{h}^{(l)}_t$ only depends on historical information, providing strict causality. We further apply residual connections and normalization to enhance gradient flow. Finally, a linear transformation with LeakyReLU activation is applied to the DCC's output $\boldsymbol{H}^{(L_{1})}$ to obtain the sequence of causal states $\tilde{\boldsymbol{S}} = (\tilde{\boldsymbol{s}}_1, \dots, \tilde{\boldsymbol{s}}_L)$:
\begin{equation}
    \tilde{\boldsymbol{S}} = \text{LeakyReLU}(\boldsymbol{W}_s \boldsymbol{H}^{(L_{1})} + \boldsymbol{B}_s),
\end{equation}
where $\boldsymbol{W}_s$ and $\boldsymbol{B}_s$ are learnable parameters.

To render the generative process ROI-constraint-aware, we introduce a constraint embedding mechanism. Specifically, the scalar Lagrange multiplier $\lambda_i$ is projected into a dense vector $\boldsymbol{e}_{\lambda_i}$ via a Multi-Layer Perceptron (MLP), serving as a global condition token: $\boldsymbol{e}_\lambda = \text{MLP}(\lambda_i)$. We prepend this constraint embedding to the causal state sequence, forming an augmented sequence $[\boldsymbol{e}_{\lambda_i}; \tilde{\boldsymbol{S}}]$.

While DCC captures local causal dependencies within a finite receptive field, we introduce a self-attention layer to model global dependencies across the entire interaction history, conditioned on $\lambda_i$. This enables the capture of long-range behavioral correlations beyond the reach of convolution. The resulting sequence is processed by self-attention followed by an FFN:
\begin{equation}
    \boldsymbol{H}_{enc}
    = \mathrm{FFN}\big(\mathrm{SelfAttn}([\boldsymbol{e}_{\lambda_i}; \tilde{\boldsymbol{S}}])\big).
\end{equation}
The output $\boldsymbol{H}_{enc}$ provides a semantic representation for the subsequent decoding stage.

% In contrast, our approach is based on autoregressive sequence modeling. However, modeling such continuous actions directly within autoregressive sequence models poses practical challenges. Existing numerical representation approaches each have specific limitations for our task: uniform bucketing~\cite{iida2021tabbie, wang2021tuta} loses precision at bucket boundaries, while the decimal decomposition~\cite{jin2021numgpt,jie2022learning} in hierarchical tokenization introduces information loss, leading to an unbalanced representation of information. Inspired by the success of Residual Quantization (RQ-VAE)~\cite{rajput2023recommender} in generative recommendation and its application to scalar modeling in generative regression tasks~\cite{ma2024generative}, we develop a structured tokenization scheme tailored for incentive allocation. 
\subsubsection{$\lambda$-Conditioned Generative Decoder}
\textbf{}
\label{subsubsec:incentive_decoding}
This module formulates incentive generation as an autoregressive sequence generation task. To overcome the precision loss inherent in mapping continuous values to discrete tokens~\cite{iida2021tabbie, wang2021tuta, jin2021numgpt}, we allow each prediction step the flexibility to choose from the vocabulary. 

Let $\boldsymbol{Z}^{(l)}$ denote the intermediate representation $\boldsymbol{H}_{enc}$ after attention mechanisms. To enforce the global ROI constraint while maintaining computational efficiency, we substitute the standard Feed-Forward Network (FFN) in the decoder with a constraint-aware Mixture-of-Experts (MoE) layer~\cite{du2022glam, zoph2022designing}. For the input vector $\boldsymbol{z} \in \boldsymbol{Z}^{(l)}$, the output $\boldsymbol{o}_{dec}$ is computed by aggregating the outputs of a sparse subset of experts:
\begin{equation}
\begin{aligned}
    \boldsymbol{o}_{\text{dec}} &= \sum_{j=1}^{N_{\text{exp}}} 
    \alpha_j(\boldsymbol{z}, \boldsymbol{c}_{\lambda}) 
    \cdot \text{Expert}_j(\boldsymbol{z}), \\
    \boldsymbol{r} &= \text{Softmax}\!\left(
    \boldsymbol{W}_g \cdot [\boldsymbol{z}; \boldsymbol{c}_{\lambda}]
    \right),\\
    \alpha_j(\boldsymbol{z}, \boldsymbol{c}_{\lambda}) &=
    \begin{cases} 
    \displaystyle \frac{r_j}{\sum_{e \in \mathcal{T}} r_e}, 
    & \begin{aligned}
    j \in \mathcal{T}, \quad
    \mathcal{T} = \text{Top-}k(\boldsymbol{r})
  \end{aligned}
 \\
    0, & \text{otherwise}.
    \end{cases}
\end{aligned}
\end{equation}

where $N_{exp}$ denotes the total number of experts, and $\alpha_j$ represents the routing weight for the $j$-th expert. In this way, the FFN is able to learn different ROI constraints and assign distinct weights to different experts (see Appendix~\ref{app:moe}).

Finally, the MoE output $\boldsymbol{o}_{dec}$ is projected into the vocabulary space to predict the next token. The generation probability for a token $\hat{g_i} \in \mathcal{V}$ is given by:
\begin{equation}
    P_\theta(\hat{g}_i^t \mid \boldsymbol{H}_{enc}, \hat{g}_i^{<t}) = \text{Softmax}(\boldsymbol{W}_{out} \cdot \boldsymbol{o}_{dec}).
\end{equation}

\noindent \textbf{Weighted Next-Token Prediction Mechanism.}
Following standard practice, we augment the vocabulary with special tokens \texttt{<SOS>}, \texttt{<EOS>}, and \texttt{<PAD>} to denote sequence boundaries and enable batch padding. These tokens carry zero incentive value.

As illustrated in Fig.~\ref{fig:framework}, the decoder generates the incentive token sequence $\hat{\boldsymbol{g}}_i = (\hat{g}_i^1, \dots, \hat{g}_i^{T_i})$ conditioned on the encoder output $\boldsymbol{H}_{enc}$ and the preceding subsequence $\hat{g}_i^{<t}$. Specifically, at generation step $t$, the output token $\hat{g}_i^t$ is computed as:
\begin{equation}
    \hat{g}_i^t = \operatorname*{arg\,max}_{g \in \mathcal{V}} P_\theta(w \mid \boldsymbol{H}_{enc}, \hat{g}_i^{<t}).
\label{eq:argmax}
\end{equation}

% Given the encoder output $\boldsymbol{H}_{\mathrm{enc}}$, the decoder produces
% a contextual representation $\boldsymbol{o}_{\mathrm{dec}}$ at each timestep.
% This representation is projected into the vocabulary space via a linear
% prediction head, yielding a categorical distribution over tokens:
% \begin{equation}
% P_\theta(g^k \mid \boldsymbol{H}_{\mathrm{enc}}, \tilde{g}^{<k})
% =
% \mathrm{Softmax}\!\left(
% \boldsymbol{W}_{\mathrm{out}} \cdot \boldsymbol{o}_{\mathrm{dec}}
% \right),
% \label{eq:ntp_prob}
% \end{equation}
% where $\boldsymbol{W}_{\mathrm{out}} \in \mathbb{R}^{|\mathcal{V}| \times d}$
% denotes the output projection matrix, and $\tilde{g}_i^{<t}$ represents
% the input token sequence provided to the decoder at timestep $t$.
To equip the model with fundamental market perception and account for the numerical significance of coin values, we employ a Position-Weighted Next Token Prediction objective. In our scenario, tokens in the early stages of the decomposition sequence represent higher orders of magnitude. Therefore, a misprediction at these positions leads to a more substantial bias in the final reward. The objective is defined as:

\begin{equation}
\mathcal{L}_{W-N T P}=-\frac{1}{N} \sum_{i=1}^N \sum_{t=1}^{T_i} \omega_t \cdot \log P_\theta\left(\hat{g}_t^i \mid \boldsymbol{H}_{enc}, \hat{g}_{<t}^i\right)
\end{equation}
where he weight $\omega_t$ is formally defined as follows: 
\begin{equation}
\omega_t=\log _{10}\left(v\left(g_t^i\right)+1\right)+\delta,
\end{equation}
where $v(\cdot)$ maps a token to its physical coin value, and $\delta$ is a smoothing constant to maintain a baseline gradient for small-value tokens. Distinct from the traditional NTP loss, which treats all token-level prediction errors indiscriminately, our weighted formulation shifts the model's focus toward numerical precision. By penalizing discrepancies based on their physical magnitude, this objective significantly refines the pre-trained model's accuracy in high-value estimation. Meanwhile, to expedite convergence, we also adopt the Teacher Forcing (TF)~\cite{venkatraman2015improving} mechanism, where the ground truth token $g_i^t$ is fed into the decoder at step $t+1$ to explicitly guide the generation process.

\subsection{Reward Model Training}
\label{subsec:RM}
Due to the extreme sparsity of final conversions, using consumption cost alone as the reward signal leads to unstable learning. To provide denser supervision, we incorporate a set of auxiliary behavioral signals $\boldsymbol{S}_{obj}$ (e.g., revenue contribution, scroll depth, retention, and task-level engagement) to enrich the reward modeling.

To aggregate these heterogeneous objectives, we adopt a neural-based fusion model to learn a personalized Preference Score (P-Score)~\cite{cao2025pantheon}. The model follows a multi-tower architecture, where each tower encodes objective-specific features, and the resulting representations are fused by a shared MLP to produce a unified utility score. The model is trained with a weighted multi-objective objective to balance trade-offs among different signals:
\begin{equation}
    \mathcal{L}_{\text{P-Score}} = \sum_{k \in \mathcal{S}_{obj}} w_k \cdot \mathcal{L}_{k}(\hat{y}, y_k^*),
\end{equation}
where $w_k$ represents the adaptive weight balancing the trade-off between objectives. For binary classification tasks (e.g., retention or click-through prediction), $\mathcal{L}_{k}$ is formulated as the standard Binary Cross-Entropy (BCE) loss:
\begin{equation}
    \mathcal{L}_{k} = - \frac{1}{N} \sum_{i=1}^{N} \left[ y_{i,k}^* \log(\hat{p}_{i,k}) + (1 - y_{i,k}^*) \log(1 - \hat{p}_{i,k}) \right],
\end{equation}
where $y_{i,k}^* \in \{0, 1\}$ denotes the ground-truth label for sample $i$ on objective $k$, and $\hat{p}_{i,k}$ is the predicted probability.

\begin{figure}[t]
  \centering
  % \vspace{0mm} % 减少图片上方与文字的间距
  \includegraphics[width=\columnwidth]{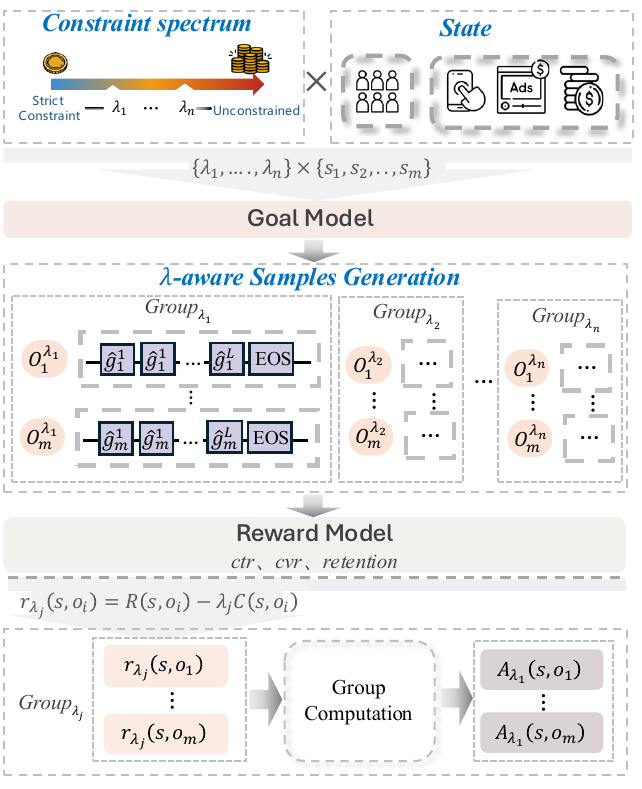}
  \caption{The workflow of Safe Constrained Policy Optimization (SCPO).}
  \label{fig:scpo}
  \vspace{-3mm} % 减少 Caption 下方与正文的间距
\end{figure}

\subsection{Safe Constrained Policy Optimization (SCPO)}
\label{subsec:SCPO}
As formulated in Section~\ref{subsec:dual}, the optimal policy for the condition-constrained alignment problem can be theoretically derived via the dual problem $\min_{\lambda \ge 0} \allowbreak \max_{\pi} \mathcal{L}(\pi, \lambda)$. However, traditional primal-dual methods require iterative updates of the multiplier $\lambda$, necessitating policy re-training whenever the ROI threshold shifts. This is computationally prohibitive for large-scale generative models.

To achieve constraint adaptation, we propose a $\lambda$-Generalization framework~\cite{zhang2021bcorle} integrated with Group Relative Policy Optimization (GRPO). We fundamentally reformulate the single-point optimization problem into a universal policy learning problem over the distribution of possible Lagrange multipliers.

\noindent \textbf{Universal Lagrangian Objective.}
Instead of treating $\lambda$ as a static hyperparameter, we elevate it to a stochastic state variable. Let $\lambda \sim P(\Lambda)$ be a multiplier sampled from a uniform distribution over a predefined range $[\lambda_{\min}, \lambda_{\max}]$. Complementary to this, the range $[\lambda_{\min}, \lambda_{\max}]$ is chosen to cover both unconstrained and strictly constrained regimes. We detail the process of determining the candidate set for $\lambda$ in Appendix~\ref{appsec:can-lambda}. We extend the state space to $\tilde{s} = [s; \lambda]$ and parameterize the policy as $\pi_\theta(a | s, \lambda)$.

The optimization objective is transformed to maximize the expected Lagrangian-conditioned reward over the entire $\lambda$ spectrum:
\begin{equation}
    \mathcal{J}_{Gen}(\theta) = \mathbb{E}_{\lambda \sim P(\Lambda)} \left[ \mathbb{E}_{\tau \sim \pi_\theta(\cdot | \lambda)} \left[ \sum_{t=0}^{T} r_\lambda(s_t, a_t) \right] \right],
\label{eq:l-obj}
\end{equation}
where, consistent with the dual formulation in Eq.~\ref{eq:dual}, the composite step-wise reward $r_\lambda$ dynamically subsumes the cost penalty based on the current $\lambda$:
\begin{equation}
    r_{\lambda}\left(s, o_i\right)=R\left(s, o_i\right)-\lambda  C\left(s, o_i\right)
\label{eq:reward}
\end{equation},
here, $R$ is the P-Score derived in Section~\ref{subsec:RM}, and $C$ represents the monetary cost of the action. This formulation incentivizes the model to learn a continuous family of strategies: aggressive revenue maximization when $\lambda \to 0$, and conservative cost control when $\lambda$ is large.

After training, the learned policy can be instantiated with different constraint levels without retraining. At inference time, a specific $\lambda^*$ is selected to satisfy the ROI constraint with the detailed selection procedure provided in Appendix~\ref{appsec:optimal_lambda}.

\noindent \textbf{\subM~with ROI-Sensitive Advantage.}
To achieve a policy adaptable to arbitrary constraint levels via preference alignment, we modify the rollout mechanism of the standard GRPO. Specifically, as illustrated in Fig.~\ref{fig:scpo}, given state $s \in \{s_1, s_2, \ldots, s_m\}$ and a set of parameters
$\{\lambda_1, \ldots, \lambda_n\}$ sampled from $[0, \lambda_{\max}]$ (A detailed formulation of the candidate selection strategy for $\lambda$ is provided in Appendix~\ref{appsec:can-lambda}.),
their Cartesian product is constructed as $\left\{\lambda_1, \ldots, \lambda_n\right\} \times\left\{s_1, s_2, \ldots, s_m\right\}$. For each pair $(\lambda_i, s_j)$, we first generate a set of candidate responses $\{o_i^{\lambda_j}\}_{i=1}^m$. Correspondingly, utilizing Eq.~\ref{eq:reward}, the reward evaluation expands from the original set $\{r_1, \dots, r_m\}$ to a constraint-conditioned matrix $\left\{r_1^{\lambda_j}\right\}, \dots, \left\{r_m^{\lambda_j}\right\}$, i.e., 
\begin{equation}
  \mathcal{R}_{\lambda_j} = \{r_{\lambda_j}(s, o_1), \dots, r_{\lambda_j}(s, o_m)\}
\end{equation}
in each Group ${\lambda_j}$. 

This augmented dataset and reward computation integrate the state, policy, and constraint variable $\lambda$, facilitating the post-training optimization of the generative policy $\pi(s, \lambda)$ to derive the optimal strategy under any ROI constraint. Finally, during the inference phase, by simply inputting the target optimal $\lambda^*$, the model generates the optimal incentive allocation strategy.

Based on the computed constraint-conditioned reward matrix $\{r^{\lambda_j}_i\}$, the subsequent step involves estimating the advantage to guide the policy gradient update. However, a critical challenge in training with variable constraint inputs is the significant scale discrepancy in rewards across different pressure levels ($\lambda$). A high penalty coefficient naturally suppresses the total reward, which could mislead the optimizer into interpreting high-$\lambda$ trajectories as poor performance solely due to the stricter constraint rather than the policy's quality. Standard global normalization fails to account for this intrinsic shift.

To address this, we define the advantage $A(s, o_i)$ using a Constraint-Conditional Normalization strategy (Appendix~\ref{app:scpo}). This approach strictly isolates the evaluation context: for a specific constraint level $\lambda_j$, we compute the advantage statistics solely over the $m$ trajectories generated under that same condition. The group statistics are computed as:
\begin{equation}
\mu_{\lambda_j} = \frac{1}{m} \sum_{i=1}^m r_{\lambda_j}(s, o_i), \quad
\sigma_{\lambda_j} = \sqrt{\frac{1}{m} \sum_{i=1}^m \left( r_{\lambda_j}(s, o_i) - \mu_{\lambda_j} \right)^2}.
\end{equation}
Consequently, the final advantage is formulated as the standardized score within its constraint group:
\begin{equation}
A(s, o_i) = \frac{r_{\lambda_j}(s, o_i) - \mu_{\lambda_j}}{\sigma_{\lambda_j} + \epsilon},
\label{eq:adv}
\end{equation}
which effectively decouples the policy's performance from the intrinsic difficulty imposed by the constraint level $\lambda$. 

The final policy optimization objective is computed by maximizing the following surrogate loss, which incorporates a PPO-style clipping mechanism and a KL-divergence regularization term:
\begin{equation}
\begin{aligned}
\mathcal{L}_{\text{SCPO}}(\theta)
&= \mathbb{E}_{s} \Bigg[
\frac{1}{mn} \sum_{i=1}^{m} \sum_{j=1}^{n}
\Big(
\min \big(
\rho_{i,j} A_{i,j}, \text{clip}(\rho_{i,j}, 1-\epsilon, 1+\epsilon) A_{i,j}
\big) \\
&- \beta \, \mathbb{D}_{\mathrm{KL}}(\pi_{\theta} \| \pi_{\text{ref}})
\Big)
\Bigg].
\end{aligned}
\label{eq:grpo_objective}
\end{equation}

\noindent where $\rho_{i,j} = \frac{\pi_{\theta}(o_i|s, \lambda_j)}{\pi_{\text{old}}(o_i|s, \lambda_j)}$ denotes the probability ratio between the current and old policies. $\epsilon$ is the clipping hyperparameter that restricts the policy update step size, and $\pi_{\text{ref}}$ represents the reference policy used to prevent excessive deviation during training.

% \paragraph{\textbf{Constraint-Aware Routing.}}
% Crucially, the effectiveness of this optimization relies on the model's ability to perceive $\lambda$. As detailed in Section~\ref{subsubsec:incentive_decoding}, the constraint explicitly guides the MoE Routing via the embedding $\boldsymbol{c}_\lambda$. The output $\boldsymbol{o}_{dec}$ is computed as:
% \begin{equation}
%     \boldsymbol{o}_{dec} = \sum_{j \in \text{Top-}k} \text{Softmax}(W_g \cdot [\boldsymbol{z}; \boldsymbol{c}_\lambda])_j \cdot \text{Expert}_j(\boldsymbol{z}).
% \end{equation}
% This structural design allows the gradients from $\mathcal{L}_{GRPO}$ to directly update the routing weights $W_g$. Consequently, the model learns to route inputs $\boldsymbol{z}$ to "conservative experts" when $\boldsymbol{c}_\lambda$ indicates a high constraint (penalizing cost) and "growth experts" when the constraint is loose, effectively encoding the Pareto frontier of the ROI-constrained allocation problem into the network parameters.

\section{Experiments}
In this section, we conduct comprehensive experiments to demonstrate the effectiveness of~\M. Specifically, we explore the following research questions:

\begin{itemize}[leftmargin=*, topsep=0pt, partopsep=0pt, parsep=0pt]
    \item \textbf{RQ1:} Does \M~achieve superior long-term revenue and ROI stability compared to state-of-the-art industrial baselines?
    \item \textbf{RQ2:} Can the proposed framework dynamically adapt to varying ROI constraints without re-training?
    \item \textbf{RQ3:} How do the key components (e.g., Causal Encoder, \subM) contribute to the overall performance?
    \item \textbf{RQ4:} How does~\M\ perform in online deployment compared to existing methods?

\end{itemize}

\subsection{Experimental Settings}
\textbf{Datasets.} 
We evaluate our method on two real-world and synthetic datasets, with the experimental setup as described below. Specifically, we use the initial six days of the data as the training set and designate the final day as the test set.
\begin{itemize}[leftmargin=*, topsep=0pt, partopsep=0pt, parsep=0pt]
    \item \textbf{IA:} This dataset is collected from the incentive advertising system of a leading short video platform, covering the period from November 5, 2025, to November 12, 2025. It comprises approximately 130,000 daily active users and 1.84 million daily incentive exposure events. 
    \item \textbf{Synthetic IA:} Constructed based on the IA dataset to maintain consistent feature alignment, this simulator introduces explicit user fatigue dynamics to benchmark long-term ROI controllability. It contains 10,000 diverse trajectories generated via a mixed behavior policy (details in Appendix~\ref{app:syn}).
\end{itemize}
\textbf{Evaluation Metrics.} We adopt a comprehensive set of metrics to assess the proposed framework from the perspectives of business growth, cost efficiency, and long-term sustainability:
\begin{itemize}[leftmargin=*, topsep=0pt, partopsep=0pt, parsep=0pt]
\item \textbf{Average Revenue (REV):} The average monetary revenue accrued by the platform from user transactions following incentive exposure.
\item \textbf{Return on Investment (ROI):} Defined as the ratio of Total Revenue to Total Incentive Cost ($\text{ROI} = \frac{\text{Revenue}}{\text{Cost}}$). This measures the economic efficiency of the allocation strategy.
\item \textbf{ROI Vialate Rate (RVR):} We propose this new metric to measure constraint satisfaction. It is defined as the percentage of sliding windows in which the realized ROI falls below the target threshold. Lower is better.
\end{itemize}

\subsection{Overall Performance (RQ1)}
\label{Sec:RQ1}
To demonstrate the comprehensive superiority of~\M, we compare it with $\lambda$=0.5 against three representative state-of-the-art approaches, covering representative paradigms of offline decision making, including generative sequence modeling methods (DT~\cite{chen2021decision} and CDT~\cite{liu2023constrained}), unconstrained offline reinforcement learning (IQL~\cite{kostrikov2021offline}) and constrained reinforcement learning approaches (CAL~\cite{wu2024off} and TREBI~\cite{lin2023safe}). More details about datasets are shown in Appendix~\ref{app:data}.

The overall performance comparisons of different approaches on the industrial dataset are reported in Table~\ref{tab:overall_performance}. From the experimental results, we can draw the following key observations: \textbf{(1) Superior Performance:} \M\ achieves a substantial improvement in total revenue while simultaneously attaining the highest ROI and a lower RVR, indicating that \M\ is able to effectively exploit incentive opportunities without sacrificing efficiency or constraint stability. \textbf{(2) Beyond MDP-Based Optimization}: The generative-based methods consistently outperform MDP-based reinforcement learning approaches across multiple evaluation metrics, clearly demonstrating the advantages of generative approaches in balancing constraint satisfaction and revenue optimization. \textbf{(3) Explicit constraint-conditioned generation is crucial for incentivized decision-making:} \M\ consistently outperforms CDT and constrained reinforcement learning methods in terms of Revenue, ROI, and RVR, indicating that explicitly conditioning the generative policy on global constraints is a more effective paradigm than post-hoc constrained optimization.

\begin{table}[ht]
  \caption{Overall performance of~\M~ on the IA dataset. The best results are shown in \textbf{bold}, and the second-best results are \underline{underlined}. Note that the REV is measured in USD.}
  \label{tab:overall_performance}
  \centering
  \resizebox{\columnwidth}{!}{
  \begin{tabular}{l|ccc|ccc}
    \toprule
    \multirow{2}{*}{\textbf{Method}} 
    & \multicolumn{3}{c|}{\textbf{IA Dataset}} 
    & \multicolumn{3}{c}{\textbf{Synthetic IA}} \\
    & \textbf{REV} $\uparrow$& \textbf{ROI} $\uparrow$& \textbf{RVR} $\downarrow$
    & \textbf{REV} $\uparrow$& \textbf{ROI} $\uparrow$& \textbf{RVR} $\downarrow$ \\
    \midrule
    DT    & \underline{0.27} & \underline{4.74} & 18.23\% & 0.19 & 3.64 & 13.51\% \\
    CDT   & 0.22 & 4.60 & \underline{15.00\%} & \textbf{0.36} & \underline{4.12} & \underline{10.24\%} \\
    IQL   & 0.25  & 3.68 &24.69\% & 0.15 & 3.59 & 18.05\% \\
    CAL   & 0.19  & 3.70 & 21.82\% & 0.12 & 3.61 & 14.49\% \\
    TREBI & 0.19 & 3.92 & 20.21\% & 0.09 & 3.72 & 14.21\% \\
    \midrule
    \textbf{GOAL (Ours)} 
          & \textbf{0.32} & \textbf{4.93} & \textbf{12.24\%} 
          & \underline{0.29} & \textbf{4.23} & \textbf{9.20\%} \\
    \emph{Improv.} 
          & \emph{+18.52\%} & \emph{+4.01\%} & \emph{+18.40\%} 
          & \emph{--} & \emph{+2.40\%} & \emph{+9.86\%} \\
    \bottomrule
  \end{tabular}
  }
  % \vspace{-3mm}
\end{table}
% \textbf{1) Dominance of GOAL:} Our proposed GOAL framework consistently achieves the best performance across all metrics and user segments. Notably, in the high-value user segment (Top 30\%), GOAL yields significant gains in GMV while maintaining a stable ROI. This demonstrates that by unifying the causal state encoder with the constraint-aware MoE mechanism, our model can effectively capture subtle fatigue signals and dynamically adjust strategies to balance aggressive revenue acquisition with strict cost compliance. \textbf{2) Limitations of Myopic Uplift Models:} While DragonNet achieves reasonable performance in short-term metrics, it significantly underperforms in long-term retention and total revenue accumulation. This confirms that conventional two-stage uplift approaches are inherently myopic, failing to account for the cumulative impact of incentives on user fatigue and future value. \textbf{3) Superiority of Sequential Modeling:} Both IQL and DT outperform the uplift baseline, particularly in the 7-day long-term evaluation. This validates the necessity of modeling incentive allocation as a sequential decision process, where capturing temporal dependencies is crucial for optimizing long-term ecosystem value.

\subsection{Dynamic Constraint Adaptation (RQ2)}

\begin{figure}[!htbp]
  \centering
  \begin{subfigure}[t]{0.48\columnwidth}
    \centering
    \includegraphics[width=\linewidth]{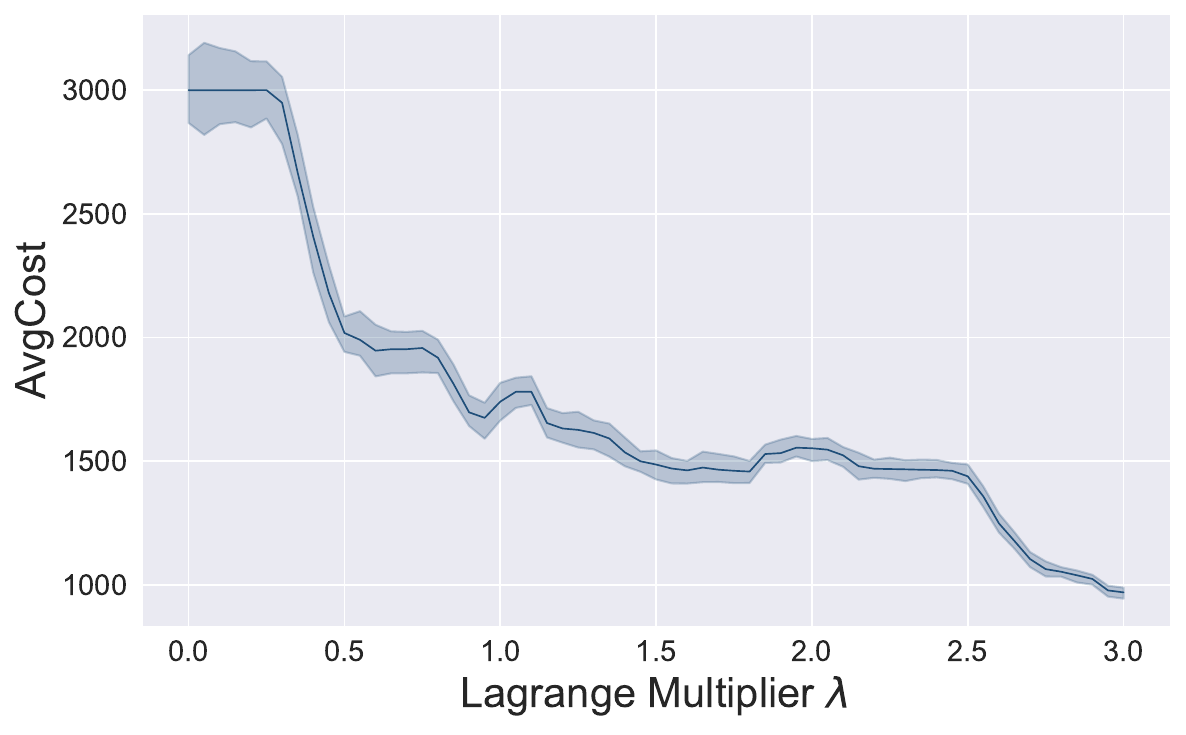}
    \caption{Average incentive cost vs. $\lambda$.}
    \label{fig:lambda_cost}
  \end{subfigure}
  \hfill
  \begin{subfigure}[t]{0.48\columnwidth}
    \centering
    \includegraphics[width=\linewidth]{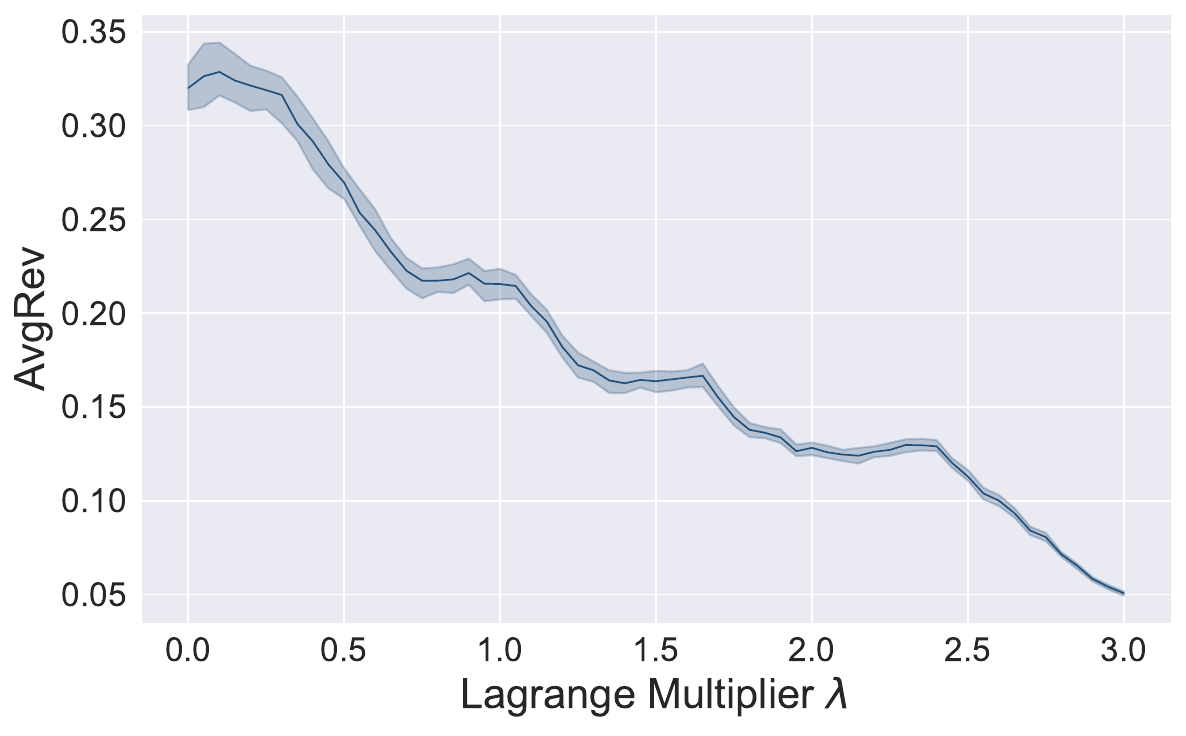}
    \caption{Average revenue vs. $\lambda$.}
    \label{fig:lambda_revenue}
  \end{subfigure}

  \caption{Effect of the Lagrangian multiplier $\lambda$ on incentive cost and revenue.}
  \label{fig:lambda_tradeoff}
\end{figure}

To answer RQ2 regarding dynamic adaptability, we conduct a sensitivity analysis on the Lagrange multiplier $\lambda$, which controls the trade-off between incentive cost and revenue.
By sweeping $\lambda$ on the test set, we examine how the learned policy adjusts its behavior under different constraint tightness levels, as shown in Fig.~\ref{fig:lambda_tradeoff}.

As illustrated in Fig.~\ref{fig:lambda_cost}, increasing $\lambda$ leads to a clear and monotonic reduction in the average incentive cost, indicating that higher constraint pressure encourages the policy to adopt more cost-efficient actions.
Meanwhile, Fig.~\ref{fig:lambda_revenue} shows that the average revenue also decreases as $\lambda$ increases, reflecting the expected trade-off between revenue maximization and cost control.

In the low-$\lambda$ regime, the model follows an aggressive strategy by allocating larger incentives to stimulate user engagement, resulting in higher cost and revenue.
As $\lambda$ increases, the policy gradually shifts toward more conservative behavior, selectively reducing incentives for low-efficiency interactions.
These results demonstrate that $\lambda$ serves as an effective inference-time control knob, enabling smooth and predictable adjustment of the policy between aggressive and conservative regimes without retraining.

\subsection{Ablation Study (RQ3)}
To assess the effectiveness of each component in~\M, we conducted a series of ablation studies. Specifically, we build several variants of~\M:

\begin{itemize}[leftmargin=*, topsep=0pt, partopsep=0pt, parsep=0pt]
    \item \textbf{w/o SCPO:} A variant of~\M\ without the Safe Constrained Policy Optimization stage, which directly uses the supervised fine-tuned (SFT) model to generate incentive tokens. 
    % \item \textbf{w/o MoE:} A variant of~\M\ without the Constraint-Aware MoE, which substitutes the MoE layer with a standard FFN.
    \item \textbf{w/o DCC:} A variant of~\M\ without the DCC module, which utilizes standard self-attention mechanisms.
    \item \textbf{SCPO w/o $\lambda$-generalizaition:} A variant of~\M\ that applies SCPO with a fixed Lagrange multiplier $\lambda$, without conditioning the policy on $\lambda$.
    \item \textbf{w/o Tokenizer:} A variant that replaces the autoregressive tokenizer with a direct regression head, outputting a single scalar instead of a token sequence under the same configurations.
\end{itemize}

\begin{table}[ht]
    \centering
    \caption{Ablation study on IA dataset. The target ROI threshold is set to $\tau = 3.0$.}
    \label{tab:ablation_study}
    \resizebox{\columnwidth}{!}{% 
    \begin{tabular}{lccc}
        \toprule
        \textbf{Method} & \textbf{REV} ($\uparrow$) & \textbf{ROI} ($\uparrow$) & \textbf{RVR} ($\downarrow$) \\
        \midrule
        w/o SCPO & 0.27 $\pm$ 0.01 & 4.57 $\pm$ 0.04 & 15.64\% $\pm$ 0.02 \\
        w/o DCC  & 0.25 $\pm$ 0.01 & 4.91 $\pm$ 0.05 & 12.29\% $\pm$ 0.02 \\
        SCPO w/o $\lambda$-gen & 0.25 $\pm$ 0.03 & 4.89 $\pm$ 0.02 & 14.12\% $\pm$ 0.04 \\
        w/o Tokenizer & 0.29 $\pm$ 0.02 & 4.89 $\pm$ 0.03 & 13.85\% $\pm$ 0.03 \\
        \midrule
        \textbf{GOAL (Ours)} & \textbf{0.32 $\pm$ 0.01} & \textbf{4.93 $\pm$ 0.02} & \textbf{12.24\% $\pm$ 0.01} \\
        \bottomrule
    \end{tabular}
    }
\end{table}

Overall, the ablation results in Table~\ref{tab:ablation_study} demonstrate that our model's effectiveness arises from the complementary roles of its components. Specifically, the "-w/o SCPO" variant leads to a pronounced degradation in ROI accompanied by an increase in CVR, indicating that the SCPO module is critical for enforcing strict ROI constraints. 
% The "-w/o MoE" variant exhibits inferior revenue and ROI performance, suggesting that the model struggles to balance the competing objectives of budget conservation and return maximization, resulting in mean-seeking behavior. 
The "-w/o DCC" variant shows substantial performance degradation across all metrics; relying on standard self-attention, the model is more prone to overfitting spurious correlations in offline data. The "-SCPO w/o $\lambda$-gen" variant achieves high ROI with low violation rates but yields the lowest revenue, indicating that a static $\lambda$ lacks the flexibility to adapt to dynamic system states. Finally, the “w/o Tokenizer” variant shows a clear drop in both REV and ROI. Without structured token generation, the model has difficulty handling the large value range and heavy-tailed distribution of incentive allocations, leading to less accurate modeling of high-stakes cases.

\subsection{Online Experimental Testing (RQ4)}

\begin{figure}[htbp!]
  \centering
  \includegraphics[width=0.88\columnwidth]{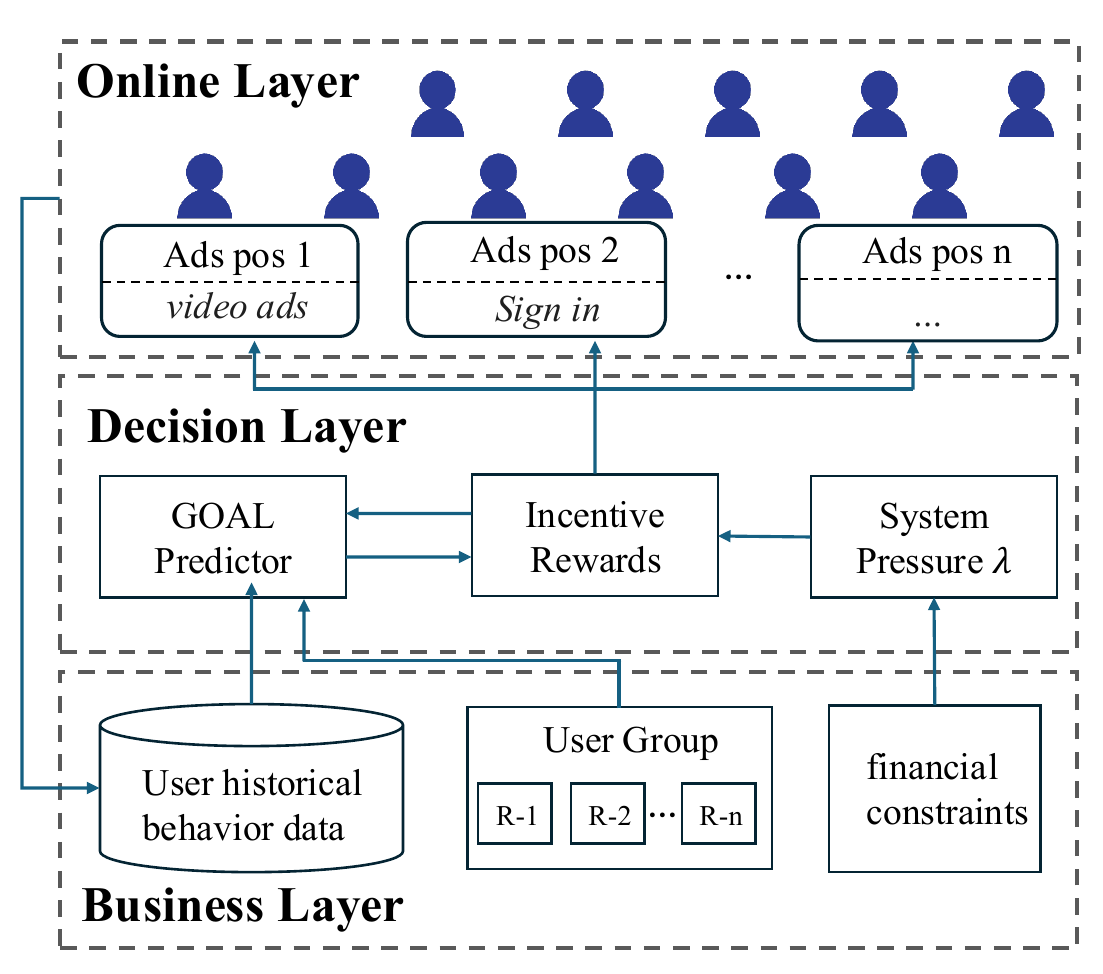}
  \caption{Online Deployment of GOAL}
  \label{fig:online}
\end{figure}

% To validate the effectiveness of our model on live traffic, we adapt our framework to the online serving environment, as shown in Fig.~\ref{fig:online}. We further conducted an A/B test on an incentivized advertising system from December 15 to December 29, 2025. Each experimental group consisted of 25\% randomly selected users. The control group used the baseline model, while the experimental group employed GOAL. Compared to the baseline, ROI increased by \textbf{2.184\%} and revenue rose by \textbf{2.559\%}. These results provide strong evidence of the effectiveness of our approach in optimizing incentivized advertising systems under strict ROI constraints.
To validate the effectiveness of our model on live traffic, we adapt our framework to the online serving environment, as shown in Fig.~\ref{fig:online}. We conducted an extended online A/B test on an incentivized advertising system spanning 4 weeks, where each experimental group encompassed 65\% of randomly selected users. The control group utilized the production baseline, while the experimental group deployed GOAL. Compared to the baseline, GOAL achieved a \textbf{2.184\%} increase in ROI and a \textbf{2.559\%} lift in revenue. Crucially, the ROI improvement remained highly stable over this longer observation window, while secondary metrics such as user dwell time and DAU exhibited no significant adverse changes. These core business gains reached statistical significance with a $p$-value of $0.03$, providing robust evidence of our approach's long-term effectiveness and financial safety in live commercial systems.

\section{Related Works}
% \subsection{Constrained Optimization in Incentive Marketing}
% \subsection{End-to-End Optimization under Constraints}
\subsection{Incentive Constraint Rewards Decision}
\label{subsec:optim}
A parallel domain is budget-constrained coupon allocation, which optimizes GMV subject to a fixed budget, distinct from our objective of maximizing revenue under strict ROI constraints. Mainstream methodologies in this domain fall into two primary categories: Resource Allocation (RA) and Direct Learning Methods(DLM). 
Approaches based on RA have evolved from disjointed two-stage paradigms~\cite{ai2022lbcf, albert2022commerce, wang2023multi} to end-to-end Decision-Focused Learning~\cite{zhou2024decision, zhang2025bi}. Works such as~\cite{ai2022lbcf, albert2022commerce, wang2023multi} typically adopt a predict-then-optimize paradigm, utilizing causal inference models to estimate average treatment effects and subsequently formulating resource allocation as a knapsack problem. In contrast,~\cite{zhou2024decision, zhang2025bi} addresses the non-differentiability of decision losses to achieve the end-to-end integration of prediction and optimization modules. Nevertheless, these approaches are limited by the difficulty of counterfactual inference, compounded by an inherent myopia that neglects future ecosystem value. 
The second line of research, DLM, bypasses the separation of ML and OR by formulating the problem as a constrained optimization problem, where a substantial body of work~\cite{xiao2019model, zhang2021bcorle, guo2025constraint} frames the coupon allocation task in sequential incentive marketing as a Constrained Markov Decision Process (CMDP) and adopts RL to solve it. These methods typically employ Lagrangian multipliers to incorporate resource constraints directly into the reward function. While capable of modeling long-term rewards, their application to incentivized advertising is hindered by the strict Markovian assumption and the inherent conservatism of offline RL algorithms~\cite{kiyohara2021accelerating, korenkevych2024offline, liu2025session}. 

Crucially, different from our business scenario, existing methods model incentives as unordered categorical labels. However, in our incentive system (e.g., coin distribution), no such categorical labels are available, making these methods not directly applicable to our precise and fine-grained continuous decision-making task.

% Crucially, beyond these algorithmic limitations, a fundamental gap exists in action space representation. Existing methods predominately treat incentives as a finite set of unordered, categorical identifiers. In contrast, incentive rewards in our context are continuous variables with strict ordinal significance (e.g., monetary magnitude). Consequently, standard allocation frameworks cannot be directly applied to the precise, fine-grained decision-making required for incentive rewards.

\subsection{Generative Sequence Modeling and Preference Alignment}
\label{subsec:overview}
The field of user behavior modeling has witnessed a fundamental paradigm shift, transitioning from discriminative matching to generative sequence modeling. Propelled by the triumph of Large Language Models (LLMs), this generative paradigm has gained significant traction. Notably, recent breakthroughs in generative recommendation~\cite{deng2025onerec, geng2022recommendation} and search~\cite{guo2025onesug, tay2022transformer} have underscored the immense potential of formulating user modeling as a sequence generation task. To steer these generative paradigms towards specific goals, researchers have adopted Preference Alignment techniques. These methods aim to calibrate the model's generative distribution to match specific reward signals. 
In pursuit of this alignment, several paradigms have emerged. RLHF \cite{ouyang2022training} typically employs PPO~\cite{schulman2017proximal} to maximize a learned reward function, yet its reliance on the Actor-Critic architecture often leads to training instability and high computational costs. To mitigate this, DPO \cite{rafailov2023direct} derives an analytical solution to implicitly optimize the reward, treating alignment as a stable classification task. Nevertheless, this approach limits the model's ability to explore beyond historical data boundaries. More recently, GRPO \cite{shao2024deepseekmath} estimates advantages via group sampling, eliminating the need for a value network to enhance efficiency. However, these methods predominantly focus on maximizing unconstrained scalar rewards, fundamentally overlooking the non-negligible cost constraints inherent to incentivized advertising.
% In this work, we represent the first attempt to formulate incentive decision-making as a sequence generation task, harnessing preference alignment to effectively model and guide complex user interaction sequences.

\section{Conclusion}
In this paper, we introduced~\M, the first generative framework designed for incentivized advertising under strict ROI constraints. By unifying causal user state modeling, constraint-aware generation, and~\subM, GOAL addresses the limitations of conventional myopic uplift models and rigid offline RL approaches. Extensive experiments on large-scale industrial datasets demonstrate that GOAL not only outperforms state-of-the-art baselines in maximizing Revenue and user retention but also exhibits superior stability in constraint adherence.
Notably, our analysis highlights the efficacy of the proposed Constraint-Aware MoE and $\lambda$-generalization mechanism, which empowers the model to adapt to dynamic operational requirements without retraining. This work marks a paradigm shift from traditional prediction-allocation pipelines to a unified generative control process, providing a scalable and flexible solution for complex decision-making in high-stakes commercial environments. In the future, we plan to extend this framework to multi-objective optimization scenarios, such as simultaneously balancing platform revenue, advertiser ROI, and user experience.

% \section*{Acknowledgment}

\bibliographystyle{ACM-Reference-Format}
\bibliography{ref}

@article{kostrikov2021offline,
  title={Offline reinforcement learning with implicit q-learning},
  author={Kostrikov, Ilya and Nair, Ashvin and Levine, Sergey},
  journal={arXiv preprint arXiv:2110.06169},
  year={2021}
}

@article{schulman2017proximal,
  title={Proximal policy optimization algorithms},
  author={Schulman, John and Wolski, Filip and Dhariwal, Prafulla and Radford, Alec and Klimov, Oleg},
  journal={arXiv preprint arXiv:1707.06347},
  year={2017}
}

@article{iida2021tabbie,
  title={Tabbie: Pretrained representations of tabular data},
  author={Iida, Hiroshi and Thai, Dung and Manjunatha, Varun and Iyyer, Mohit},
  journal={arXiv preprint arXiv:2105.02584},
  year={2021}
}

@article{jin2021numgpt,
  title={Numgpt: Improving numeracy ability of generative pre-trained models},
  author={Jin, Zhihua and Jiang, Xin and Wang, Xingbo and Liu, Qun and Wang, Yong and Ren, Xiaozhe and Qu, Huamin},
  journal={arXiv preprint arXiv:2109.03137},
  year={2021}
}

@inproceedings{wang2021tuta,
  title={Tuta: Tree-based transformers for generally structured table pre-training},
  author={Wang, Zhiruo and Dong, Haoyu and Jia, Ran and Li, Jia and Fu, Zhiyi and Han, Shi and Zhang, Dongmei},
  booktitle={Proceedings of the 27th ACM SIGKDD Conference on Knowledge Discovery \& Data Mining},
  pages={1780--1790},
  year={2021}
}

@inproceedings{jie2022learning,
  title={Learning to Reason Deductively: Math Word Problem Solving as Complex Relation Extraction},
  author={Jie, Zhanming and Li, Jierui and Lu, Wei},
  booktitle={Proceedings of the 60th Annual Meeting of the Association for Computational Linguistics (Volume 1: Long Papers)},
  pages={5944--5955},
  year={2022}
}

@article{rajput2023recommender,
  title={Recommender systems with generative retrieval},
  author={Rajput, Shashank and Mehta, Nikhil and Singh, Anima and Hulikal Keshavan, Raghunandan and Vu, Trung and Heldt, Lukasz and Hong, Lichan and Tay, Yi and Tran, Vinh and Samost, Jonah and others},
  journal={Advances in Neural Information Processing Systems},
  volume={36},
  pages={10299--10315},
  year={2023}
}

@article{ma2024generative,
  title={Generative Regression Based Watch Time Prediction for Short-Video Recommendation},
  author={Ma, Hongxu and Tian, Kai and Zhang, Tao and Zhang, Xuefeng and Zhou, Han and Chen, Chunjie and Li, Han and Guan, Jihong and Zhou, Shuigeng},
  journal={arXiv preprint arXiv:2412.20211},
  year={2024}
}

@inproceedings{goldenberg2020free,
  title={Free lunch! retrospective uplift modeling for dynamic promotions recommendation within roi constraints},
  author={Goldenberg, Dmitri and Albert, Javier and Bernardi, Lucas and Estevez, Pablo},
  booktitle={Proceedings of the 14th ACM Conference on Recommender Systems},
  pages={486--491},
  year={2020}
}

@inproceedings{wu2018budget,
  title={Budget constrained bidding by model-free reinforcement learning in display advertising},
  author={Wu, Di and Chen, Xiujun and Yang, Xun and Wang, Hao and Tan, Qing and Zhang, Xiaoxun and Xu, Jian and Gai, Kun},
  booktitle={Proceedings of the 27th ACM International Conference on Information and Knowledge Management},
  pages={1443--1451},
  year={2018}
}

@article{hansotia2002direct,
  title={Direct marketing for multichannel retailers: Issues, challenges and solutions},
  author={Hansotia, Behram J and Rukstales, Bradley},
  journal={Journal of Database Marketing \& Customer Strategy Management},
  volume={9},
  number={3},
  pages={259--266},
  year={2002},
  publisher={Springer}
}

@inproceedings{zhao2019unified,
  title={A unified framework for marketing budget allocation},
  author={Zhao, Kui and Hua, Junhao and Yan, Ling and Zhang, Qi and Xu, Huan and Yang, Cheng},
  booktitle={Proceedings of the 25th ACM SIGKDD International Conference on Knowledge Discovery \& Data Mining},
  pages={1820--1830},
  year={2019}
}

@inproceedings{pei2019value,
  title={Value-aware recommendation based on reinforcement profit maximization},
  author={Pei, Changhua and Yang, Xinru and Cui, Qing and Lin, Xiao and Sun, Fei and Jiang, Peng and Ou, Wenwu and Zhang, Yongfeng},
  booktitle={The World Wide Web Conference},
  pages={3123--3129},
  year={2019}
}

@inproceedings{zou2019reinforcement,
  title={Reinforcement learning to optimize long-term user engagement in recommender systems},
  author={Zou, Lixin and Xia, Long and Ding, Zhuoye and Song, Jiaxing and Liu, Weidong and Yin, Dawei},
  booktitle={Proceedings of the 25th ACM SIGKDD international conference on knowledge discovery \& data mining},
  pages={2810--2818},
  year={2019}
}

@inproceedings{zheng2018drn,
  title={DRN: A deep reinforcement learning framework for news recommendation},
  author={Zheng, Guanjie and Zhang, Fuzheng and Zheng, Zihan and Xiang, Yang and Yuan, Nicholas Jing and Xie, Xing and Li, Zhenhui},
  booktitle={Proceedings of the 2018 world wide web conference},
  pages={167--176},
  year={2018}
}

@article{zhang2021bcorle,
  title={BCORLE ($\lambda$): An Offline Reinforcement Learning and Evaluation Framework for Coupons Allocation in E-commerce Market},
  author={Zhang, Yang and Tang, Bo and Yang, Qingyu and An, Dou and Tang, Hongyin and Xi, Chenyang and Li, Xueying and Xiong, Feiyu},
  journal={Advances in Neural Information Processing Systems},
  volume={34},
  pages={20410--20422},
  year={2021}
}

@article{chen2022bcrlsp,
  title={BCRLSP: An offline reinforcement learning framework for sequential targeted promotion},
  author={Chen, Fanglin and Liu, Xiao and Tang, Bo and Xiong, Feiyu and Hwang, Serim and Zhuang, Guomian},
  journal={arXiv preprint arXiv:2207.07790},
  year={2022}
}

@article{kiyohara2021accelerating,
  title={Accelerating offline reinforcement learning application in real-time bidding and recommendation: Potential use of simulation},
  author={Kiyohara, Haruka and Kawakami, Kosuke and Saito, Yuta},
  journal={arXiv preprint arXiv:2109.08331},
  year={2021}
}

@inproceedings{korenkevych2024offline,
  title={Offline reinforcement learning for optimizing production bidding policies},
  author={Korenkevych, Dmytro and Cheng, Frank and Balakir, Artsiom and Nikulkov, Alex and Gao, Lingnan and Cen, Zhihao and Xu, Zuobing and Zhu, Zheqing},
  booktitle={Proceedings of the 30th ACM SIGKDD Conference on Knowledge Discovery and Data Mining},
  pages={5251--5259},
  year={2024}
}

@article{liu2025session,
  title={Session-Level Dynamic Ad Load Optimization using Offline Robust Reinforcement Learning},
  author={Liu, Tao and Xu, Qi and Shi, Wei and Hua, Zhigang and Yang, Shuang},
  journal={arXiv preprint arXiv:2501.05591},
  year={2025}
}

@article{li2023survey,
  title={A survey on transformers in reinforcement learning},
  author={Li, Wenzhe and Luo, Hao and Lin, Zichuan and Zhang, Chongjie and Lu, Zongqing and Ye, Deheng},
  journal={arXiv preprint arXiv:2301.03044},
  year={2023}
}

@article{xiao2021early,
  title={Early convolutions help transformers see better},
  author={Xiao, Tete and Singh, Mannat and Mintun, Eric and Darrell, Trevor and Doll{\'a}r, Piotr and Girshick, Ross},
  journal={Advances in neural information processing systems},
  volume={34},
  pages={30392--30400},
  year={2021}
}

@inproceedings{geng2022recommendation,
  title={Recommendation as language processing (rlp): A unified pretrain, personalized prompt \& predict paradigm (p5)},
  author={Geng, Shijie and Liu, Shuchang and Fu, Zuohui and Ge, Yingqiang and Zhang, Yongfeng},
  booktitle={Proceedings of the 16th ACM conference on recommender systems},
  pages={299--315},
  year={2022}
}

@article{deng2025onerec,
  title={Onerec: Unifying retrieve and rank with generative recommender and iterative preference alignment},
  author={Deng, Jiaxin and Wang, Shiyao and Cai, Kuo and Ren, Lejian and Hu, Qigen and Ding, Weifeng and Luo, Qiang and Zhou, Guorui},
  journal={arXiv preprint arXiv:2502.18965},
  year={2025}
}

@inproceedings{du2022glam,
  title={Glam: Efficient scaling of language models with mixture-of-experts},
  author={Du, Nan and Huang, Yanping and Dai, Andrew M and Tong, Simon and Lepikhin, Dmitry and Xu, Yuanzhong and Krikun, Maxim and Zhou, Yanqi and Yu, Adams Wei and Firat, Orhan and others},
  booktitle={International conference on machine learning},
  pages={5547--5569},
  year={2022},
  organization={PMLR}
}

@article{shao2024deepseekmath,
  title={Deepseekmath: Pushing the limits of mathematical reasoning in open language models},
  author={Shao, Zhihong and Wang, Peiyi and Zhu, Qihao and Xu, Runxin and Song, Junxiao and Bi, Xiao and Zhang, Haowei and Zhang, Mingchuan and Li, YK and Wu, Yang and others},
  journal={arXiv preprint arXiv:2402.03300},
  year={2024}
}

@inproceedings{hausknecht2015deep,
  title={Deep Recurrent Q-Learning for Partially Observable MDPs.},
  author={Hausknecht, Matthew J and Stone, Peter},
  booktitle={AAAI fall symposia},
  volume={45},
  pages={141},
  year={2015}
}

@inproceedings{zhao2018deep,
  title={Deep reinforcement learning for page-wise recommendations},
  author={Zhao, Xiangyu and Xia, Long and Zhang, Liang and Ding, Zhuoye and Yin, Dawei and Tang, Jiliang},
  booktitle={Proceedings of the 12th ACM conference on recommender systems},
  pages={95--103},
  year={2018}
}

@inproceedings{li2024modeling,
  title={Modeling user fatigue for sequential recommendation},
  author={Li, Nian and Ban, Xin and Ling, Cheng and Gao, Chen and Hu, Lantao and Jiang, Peng and Gai, Kun and Li, Yong and Liao, Qingmin},
  booktitle={Proceedings of the 47th International ACM SIGIR Conference on Research and Development in Information Retrieval},
  pages={996--1005},
  year={2024}
}

@inproceedings{zhou2021informer,
  title={Informer: Beyond efficient transformer for long sequence time-series forecasting},
  author={Zhou, Haoyi and Zhang, Shanghang and Peng, Jieqi and Zhang, Shuai and Li, Jianxin and Xiong, Hui and Zhang, Wancai},
  booktitle={Proceedings of the AAAI conference on artificial intelligence},
  volume={35},
  number={12},
  pages={11106--11115},
  year={2021}
}

@article{liu2024lost,
  title={Lost in the middle: How language models use long contexts},
  author={Liu, Nelson F and Lin, Kevin and Hewitt, John and Paranjape, Ashwin and Bevilacqua, Michele and Petroni, Fabio and Liang, Percy},
  journal={Transactions of the Association for Computational Linguistics},
  volume={12},
  pages={157--173},
  year={2024}
}

@article{ouyang2022training,
  title={Training language models to follow instructions with human feedback},
  author={Ouyang, Long and Wu, Jeffrey and Jiang, Xu and Almeida, Diogo and Wainwright, Carroll and Mishkin, Pamela and Zhang, Chong and Agarwal, Sandhini and Slama, Katarina and Ray, Alex and others},
  journal={Advances in neural information processing systems},
  volume={35},
  pages={27730--27744},
  year={2022}
}

@article{rafailov2023direct,
  title={Direct preference optimization: Your language model is secretly a reward model},
  author={Rafailov, Rafael and Sharma, Archit and Mitchell, Eric and Manning, Christopher D and Ermon, Stefano and Finn, Chelsea},
  journal={Advances in neural information processing systems},
  volume={36},
  pages={53728--53741},
  year={2023}
}

@article{guo2025onesug,
  title={OneSug: The Unified End-to-End Generative Framework for E-commerce Query Suggestion},
  author={Guo, Xian and Chen, Ben and Wang, Siyuan and Yang, Ying and Lei, Chenyi and Ding, Yuqing and Li, Han},
  journal={arXiv preprint arXiv:2506.06913},
  year={2025}
}

@article{tay2022transformer,
  title={Transformer memory as a differentiable search index},
  author={Tay, Yi and Tran, Vinh and Dehghani, Mostafa and Ni, Jianmo and Bahri, Dara and Mehta, Harsh and Qin, Zhen and Hui, Kai and Zhao, Zhe and Gupta, Jai and others},
  journal={Advances in Neural Information Processing Systems},
  volume={35},
  pages={21831--21843},
  year={2022}
}

@inproceedings{ai2022lbcf,
  title={Lbcf: A large-scale budget-constrained causal forest algorithm},
  author={Ai, Meng and Li, Biao and Gong, Heyang and Yu, Qingwei and Xue, Shengjie and Zhang, Yuan and Zhang, Yunzhou and Jiang, Peng},
  booktitle={Proceedings of the ACM Web Conference 2022},
  pages={2310--2319},
  year={2022}
}

@inproceedings{albert2022commerce,
  title={E-commerce promotions personalization via online multiple-choice knapsack with uplift modeling},
  author={Albert, Javier and Goldenberg, Dmitri},
  booktitle={Proceedings of the 31st ACM International Conference on Information \& Knowledge Management},
  pages={2863--2872},
  year={2022}
}

@inproceedings{wang2023multi,
  title={A Multi-stage Framework for Online Bonus Allocation Based on Constrained User Intent Detection},
  author={Wang, Chao and Shi, Xiaowei and Xu, Shuai and Wang, Zhe and Fan, Zhiqiang and Feng, Yan and You, An and Chen, Yu},
  booktitle={Proceedings of the 29th ACM SIGKDD Conference on Knowledge Discovery and Data Mining},
  pages={5028--5038},
  year={2023}
}

@inproceedings{zhou2024decision,
  title={Decision focused causal learning for direct counterfactual marketing optimization},
  author={Zhou, Hao and Huang, Rongxiao and Li, Shaoming and Jiang, Guibin and Zheng, Jiaqi and Cheng, Bing and Lin, Wei},
  booktitle={Proceedings of the 30th ACM SIGKDD Conference on Knowledge Discovery and Data Mining},
  pages={6368--6379},
  year={2024}
}

@article{zhang2025bi,
  title={Bi-Level Decision-Focused Causal Learning for Large-Scale Marketing Optimization: Bridging Observational and Experimental Data},
  author={Zhang, Shuli and Zhou, Hao and Zheng, Jiaqi and Jiang, Guibin and Cheng, Bing and Lin, Wei and Chen, Guihai},
  journal={arXiv preprint arXiv:2510.19517},
  year={2025}
}

@inproceedings{xiao2019model,
  title={Model-based constrained MDP for budget allocation in sequential incentive marketing},
  author={Xiao, Shuai and Guo, Le and Jiang, Zaifan and Lv, Lei and Chen, Yuanbo and Zhu, Jun and Yang, Shuang},
  booktitle={Proceedings of the 28th ACM International Conference on Information and Knowledge Management},
  pages={971--980},
  year={2019}
}

@inproceedings{guo2025constraint,
  title={Constraint-conditioned actor-critic for offline safe reinforcement learning},
  author={Guo, Zijian and Zhou, Weichao and Wang, Shengao and Li, Wenchao},
  booktitle={The Thirteenth International Conference on Learning Representations},
  year={2025}
}

@article{bai2018empirical,
  title={An Empirical Evaluation of Generic Convolutional and Recurrent Networks for Sequence Modeling},
  author={Bai, Shaojie},
  journal={arXiv preprint arXiv:1803.01271},
  year={2018}
}

@inproceedings{cao2025pantheon,
  title={Pantheon: Personalized multi-objective ensemble sort via iterative pareto policy optimization},
  author={Cao, Jiangxia and Xu, Pengbo and Cheng, Yin and Guo, Kaiwei and Tang, Jian and Wang, Shijun and Leng, Dewei and Yang, Shuang and Liu, Zhaojie and Niu, Yanan and others},
  booktitle={Proceedings of the 34th ACM International Conference on Information and Knowledge Management},
  pages={5575--5582},
  year={2025}
}

@article{wallace2019nlp,
  title={Do NLP models know numbers? probing numeracy in embeddings},
  author={Wallace, Eric and Wang, Yizhong and Li, Sujian and Singh, Sameer and Gardner, Matt},
  journal={arXiv preprint arXiv:1909.07940},
  year={2019}
}

@inproceedings{venkatraman2015improving,
  title={Improving multi-step prediction of learned time series models},
  author={Venkatraman, Arun and Hebert, Martial and Bagnell, J},
  booktitle={Proceedings of the AAAI Conference on Artificial Intelligence},
  volume={29},
  number={1},
  year={2015}
}

@article{chen2021decision,
  title={Decision transformer: Reinforcement learning via sequence modeling},
  author={Chen, Lili and Lu, Kevin and Rajeswaran, Aravind and Lee, Kimin and Grover, Aditya and Laskin, Misha and Abbeel, Pieter and Srinivas, Aravind and Mordatch, Igor},
  journal={Advances in neural information processing systems},
  volume={34},
  pages={15084--15097},
  year={2021}
}

@inproceedings{liu2023constrained,
  title={Constrained decision transformer for offline safe reinforcement learning},
  author={Liu, Zuxin and Guo, Zijian and Yao, Yihang and Cen, Zhepeng and Yu, Wenhao and Zhang, Tingnan and Zhao, Ding},
  booktitle={International conference on machine learning},
  pages={21611--21630},
  year={2023},
  organization={PMLR}
}

@article{wu2024off,
  title={Off-policy primal-dual safe reinforcement learning},
  author={Wu, Zifan and Tang, Bo and Lin, Qian and Yu, Chao and Mao, Shangqin and Xie, Qianlong and Wang, Xingxing and Wang, Dong},
  journal={arXiv preprint arXiv:2401.14758},
  year={2024}
}

@inproceedings{lin2023safe,
  title={Safe offline reinforcement learning with real-time budget constraints},
  author={Lin, Qian and Tang, Bo and Wu, Zifan and Yu, Chao and Mao, Shangqin and Xie, Qianlong and Wang, Xingxing and Wang, Dong},
  booktitle={International Conference on Machine Learning},
  pages={21127--21152},
  year={2023},
  organization={PMLR}
}

@inproceedings{zoph2022designing,
  title={Designing effective sparse expert models},
  author={Zoph, Barret},
  booktitle={2022 IEEE International Parallel and Distributed Processing Symposium Workshops (IPDPSW)},
  pages={1044--1044},
  year={2022},
  organization={IEEE}
}

\appendix
\section*{APPENDIX}

%%%%%%%%%%%%%%%%%%%%%%%%%%%%%%%%%%%%%%%%%%%
\section{Selection of Candidate $\lambda$ Sets}
\label{appsec:can-lambda}
%%%%%%%%%%%%%%%%%%%%%%%%%%%%%%%%%%%%%%%%%%%

To enable the policy to perceive and adapt to varying business requirements without retraining, we formalize a Goal-conditioned Policy $\pi(a|s, \lambda)$. In this framework, the state space is augmented with the Lagrangian multiplier $\lambda$ as a contextual indicator. The objective is to learn a universal strategy manifold that approximates the Pareto frontier between revenue and ROI.

A fundamental challenge is to define an effective sampling interval $[\lambda_{\min}, \lambda_{\max}]$ that covers all potentially optimal strategies. If $\lambda$ is too small, the policy fails to satisfy ROI constraints; if $\lambda$ is excessively large, the policy may collapse prematurely. We derive the boundaries based on the marginal utility of actions:

\begin{itemize}[leftmargin=*]
    \item \textbf{Lower Bound ($\lambda_{\min} = 0$):} When $\lambda = 0$, the objective $r_i - \lambda c_i$ reduces to pure revenue maximization. This represents the Unconstrained Boundary, where the policy explores the maximum possible incentive intensity to establish the upper limit of revenue.

    \item \textbf{Upper Bound ($\lambda_{\max} = \tau$):}  According to the derived dual relationship $\lambda = \frac{\lambda' \tau}{1 + \lambda'}$ in Eq.~\ref{eq:dual}, the effective penalty coefficient $\lambda$ asymptotically approaches the ROI threshold $\tau$ as the dual penalty $\lambda' \to \infty$. 
\end{itemize}

To ensure the theoretical rigor of this boundary, we introduce the following property:

\noindent \textbf{Proposition 1}. For a discrete action set $\mathcal{A}$ with ascending costs $c_1 < c_2 < \dots < c_k$, there exists a finite threshold $\bar{\lambda} = \max_{i>1} \frac{r_i - r_1}{c_i - c_1}$ such that for all $\lambda \ge \bar{\lambda}$, the optimal action $a = \arg\max_{a_i \in \mathcal{A}} (r_i - \lambda c_i)$ consistently converges to the minimum-cost action $a_1$.

% \textit{Proof.} Consider the objective function $\mathcal{F}(\lambda, a_i) = r_i - \lambda c_i$. We aim to find the condition where $a_1$ is the unique optimizer, i.e., $a_1 = \arg\max_{a_i} \mathcal{F}(\lambda, a_i)$, which must satisfy the following system of linear inequalities:$$r_1 - \lambda c_1 > r_i - \lambda c_i, \quad \forall i \in \{2, \dots, k\}$$Isolating the terms involving $\lambda$ yields:$$\lambda(c_i - c_1) > r_i - r_1$$Given the premise of ascending costs $c_i > c_j$ for $i > j$, the coefficient $(c_i - c_1)$ is strictly positive for all $i > 1$. Dividing both sides by this term defines a set of lower bounds for the multiplier $\lambda$:$$\lambda > \frac{r_i - r_1}{c_i - c_1}, \quad \forall i \in \{2, \dots, k\}$$We define the critical threshold $\bar{\lambda}$ as the supremum of these individual action-wise bounds:$$\bar{\lambda} = \max_{i \in \{2, \dots, k\}} \left( \frac{r_i - r_1}{c_i - c_1} \right)$$It follows that for any $\lambda \ge \bar{\lambda}$, the inequality $\mathcal{F}(\lambda, a_1) \ge \mathcal{F}(\lambda, a_i)$ is guaranteed for the entire action set $\mathcal{A}$. In this regime, the marginal penalty imposed by $\lambda$ per unit of cost exceeds the marginal reward gain of any alternative action $a_i$ relative to $a_1$. Thus, the optimal policy $\pi^*$ necessarily converges to the point mass on $a_1$. This completes the proof. $\square$

\textit{Proof.} Consider the objective function $\mathcal{F}(\lambda, a_i) = r_i - \lambda c_i$. We aim to find the condition where $a_1$ is the unique optimizer, i.e., $a_1 = \arg\max_{a_i} \mathcal{F}(\lambda, a_i)$, which must satisfy the system of linear inequalities $r_1 - \lambda c_1 > r_i - \lambda c_i$ for all $i \in \{2, \dots, k\}$. Isolating the terms involving $\lambda$ yields $\lambda(c_i - c_1) > r_i - r_1$. Given the premise of ascending costs $c_i > c_j$ for $i > j$, the coefficient $(c_i - c_1)$ is strictly positive for all $i > 1$. Dividing both sides by this term defines a set of lower bounds for the multiplier $\lambda$ as $\lambda > \frac{r_i - r_1}{c_i - c_1}$ for all $i \in \{2, \dots, k\}$. We define the critical threshold $\bar{\lambda}$ as the supremum of these individual action-wise bounds: $\bar{\lambda} = \max_{i \in \{2, \dots, k\}} \left( \frac{r_i - r_1}{c_i - c_1} \right)$. It follows that for any $\lambda \ge \bar{\lambda}$, the inequality $\mathcal{F}(\lambda, a_1) \ge \mathcal{F}(\lambda, a_i)$ is guaranteed for the entire action set $\mathcal{A}$. In this regime, the marginal penalty imposed by $\lambda$ per unit of cost exceeds the marginal reward gain of any alternative action $a_i$ relative to $a_1$. Thus, the optimal policy $\pi^*$ necessarily converges to the point mass on $a_1$. This completes the proof. $\square$

In practical scenarios, business requirements often specify a minimum ROI threshold $\tau$ that varies across different marketing campaigns. To ensure our policy can generalize to the most stringent constraints, we identify the maximum value among these potential ROI targets, denoted as $\tau_{\max}$. Since $\tau_{\max}$ typically exceeds the marginal utility threshold $\bar{\lambda}$, we utilize it as the practical sampling ceiling $\lambda_{\max}$. This ensures that the training interval $[0, \lambda_{\max}]$ encompasses all operationally relevant strategy transitions, from unconstrained growth to the most conservative efficiency requirements.

To implement this in our training pipeline, we adopt a discretization and uniform sampling approach. Specifically, the continuous interval $[0, \lambda_{\max}]$ is partitioned into a candidate set $\Lambda = \{0, \Delta\lambda, 2\Delta\lambda, \dots, \lambda_{\max}\}$ with a fixed step size $\Delta\lambda$. For instance, with $\lambda_{\max} = 3.0$ and $\Delta\lambda = 0.5$, the model explores 7 distinct cost-penalty scenarios. During each training iteration, a target $\lambda$ is randomly drawn from $\Lambda$ and concatenated with the system state $s$, enabling the policy $\pi(a|s, \lambda)$ to internalize the mapping between varying ROI constraints and their corresponding optimal decision logics. This stochastic exposure ensures the model approximates a continuous Pareto frontier rather than over-fitting to a single trade-off point.

\section{Selection of Optimal $\lambda^*$ at Inference}
\label{appsec:optimal_lambda}

To ensure the learned policy strictly adheres to the target ROI threshold $\tau$ during deployment, we employ a binary search procedure to identify the optimal dual variable $\lambda^*$, leveraging the monotonicity of the realized ROI with respect to $\lambda$, as detailed in Algorithm~\ref{alg:calibration}.

\begin{algorithm}
\caption{$\lambda$ Calibration}
\label{alg:calibration}
\begin{algorithmic}[1]
\REQUIRE Trained Policy $\pi_\theta$, Calibration Dataset $\mathcal{D}$, Target ROI threshold $\tau$
\REQUIRE Constraint spectrum $\Lambda_{cand} = \{\lambda_1, \dots, \lambda_n\}$ (sorted in ascending order)
\REQUIRE Tolerance $\epsilon$ 

\STATE Initialize: $low \leftarrow 1, \ high \leftarrow n, \ \lambda^* \leftarrow \lambda_n$ \textcolor[RGB]{0, 0, 128}{\COMMENT{\textit{Default to safety}}}

\WHILE{$low \le high$}
    \STATE $mid \leftarrow \lfloor (low + high) / 2 \rfloor$
    \STATE $\lambda_{curr} \leftarrow \Lambda_{cand}[mid]$
    
    \STATE \textcolor[RGB]{0, 0, 128}{\COMMENT{\textit{Empirical Policy Evaluation}}}
    \STATE Rollout: Generate actions $a \sim \pi_\theta(\cdot|s, \lambda_{curr})$ for all $s \in \mathcal{D}$
    \STATE Calculate Realized ROI: $\hat{\mathcal{R}} \leftarrow \frac{\sum r_i}{\sum c_i}$
    
    \STATE \textcolor[RGB]{0, 0, 128}{\COMMENT{\textit{Binary Search Range Update}}}
    \IF{$\hat{\mathcal{R}} \ge \tau$}
        \STATE $\lambda^* \leftarrow \lambda_{curr}$ \textcolor[RGB]{0, 0, 128}{\COMMENT{\textit{ROI threshold satisfied}}}
        \STATE $high \leftarrow mid - 1$
    \ELSE
        \STATE $low \leftarrow mid + 1$ \textcolor[RGB]{0, 0, 128}{\COMMENT{\textit{ROI threshold violated}}}
    \ENDIF
\ENDWHILE

\RETURN $\lambda^*$
\end{algorithmic}
\end{algorithm}

\section{Safe Constrained Policy Optimization}
\label{app:scpo}
The algorithmic details of SCPO are presented in Algorith~\ref{alg:scpo}.
\begin{algorithm}[t]
\caption{Safe Constrained Policy Optimization (SCPO)}
\label{alg:scpo}
\begin{algorithmic}[1]
\REQUIRE User state dataset $\mathcal{D}$, Constraint spectrum $\Lambda = \{\lambda_1, \dots, \lambda_n\}$
\REQUIRE Policy $\pi_\theta$, Reference policy $\pi_{\text{ref}}$, Group size $m$
\REQUIRE Hyperparameters: clipping $\epsilon$, KL coefficient $\beta$, Learning rate $\eta$
\WHILE{not converged}
    \STATE Sample a batch of states $B_s \sim \mathcal{D}$
    \FOR{each state $s \in B_s$}
        \STATE \textcolor[RGB]{0, 0, 128}{\COMMENT{\textit{Candidate Generation}}}
        \STATE Sample $m$ candidate responses $\{o_i\}_{i=1}^m \sim \pi_{\theta}(\cdot|s)$ 
        \STATE \textcolor[RGB]{0, 0, 128}{\COMMENT{\textit{Cartesian Augmentation \& Reward Evaluation}}}
        \FOR{each constraint level $\lambda_j \in \Lambda$}
            \STATE Compute rewards for all candidates under the condition $\lambda_j$:
            \STATE $r_{i,j} \leftarrow R(s, o_i) - \lambda_j C(s, o_i), \quad \forall i \in \{1, \dots, m\}$
            
            \STATE \textcolor[RGB]{0, 0, 128}{\COMMENT{\textit{Constraint-Conditional Normalization}}}
            \STATE Compute group statistics specifically for $\lambda_j$:
            \STATE $\mu_{\lambda_j} \leftarrow \frac{1}{m} \sum_{i=1}^m r_{i,j}, \quad \sigma_{\lambda_j} \leftarrow \sqrt{\frac{1}{m} \sum_{i=1}^m (r_{i,j} - \mu_{\lambda_j})^2}$
            
            \STATE Compute normalized advantages (Eq.~\ref{eq:adv}):
            \STATE $A_{i,j} \leftarrow \frac{r_{i,j} - \mu_{\lambda_j}}{\sigma_{\lambda_j} + \epsilon}, \quad \forall i \in \{1, \dots, m\}$
        \ENDFOR
    \ENDFOR
    
    \STATE \textcolor[RGB]{0, 0, 128}{\COMMENT{\textit{Policy Optimization}}}
    \STATE Construct the surrogate objective $\mathcal{L}_{\text{SCPO}}(\theta)$ over the augmented batch:
    \STATE $
    \begin{aligned}
    \mathcal{L} \leftarrow &
    \frac{1}{|B_s|\, m\, n}
    \sum_{s} \sum_{i=1}^m \sum_{j=1}^n
    \Big[
    \min\!\left(
    \rho_{i,j} A_{i,j},
    \operatorname{clip}\!\left(
    \rho_{i,j},
    1-\epsilon,
    \right. \right.
    \\
    &\left. \left.
    1+\epsilon
    \right) A_{i,j}
    \right)
    - \beta D_{\mathrm{KL}}
    \Big]
    \end{aligned}
    $
    \STATE Update policy parameters: $\theta \leftarrow \theta + \eta \nabla_\theta \mathcal{L}$
\ENDWHILE
\end{algorithmic}
\end{algorithm}

\section{Constraint-aware Experts}
\label{app:moe}
To understand how the constraint-aware MoE internalizes the global constraint signal, we analyze the routing behavior of the gating network under different values of the Lagrange multiplier $\lambda$.
Recall that $\lambda$ explicitly parameterizes the strength of the ROI constraint and is provided to the gating network as a global conditioning signal.

\begin{figure}[!htbp]
  \centering
  \includegraphics[width=0.9\columnwidth]{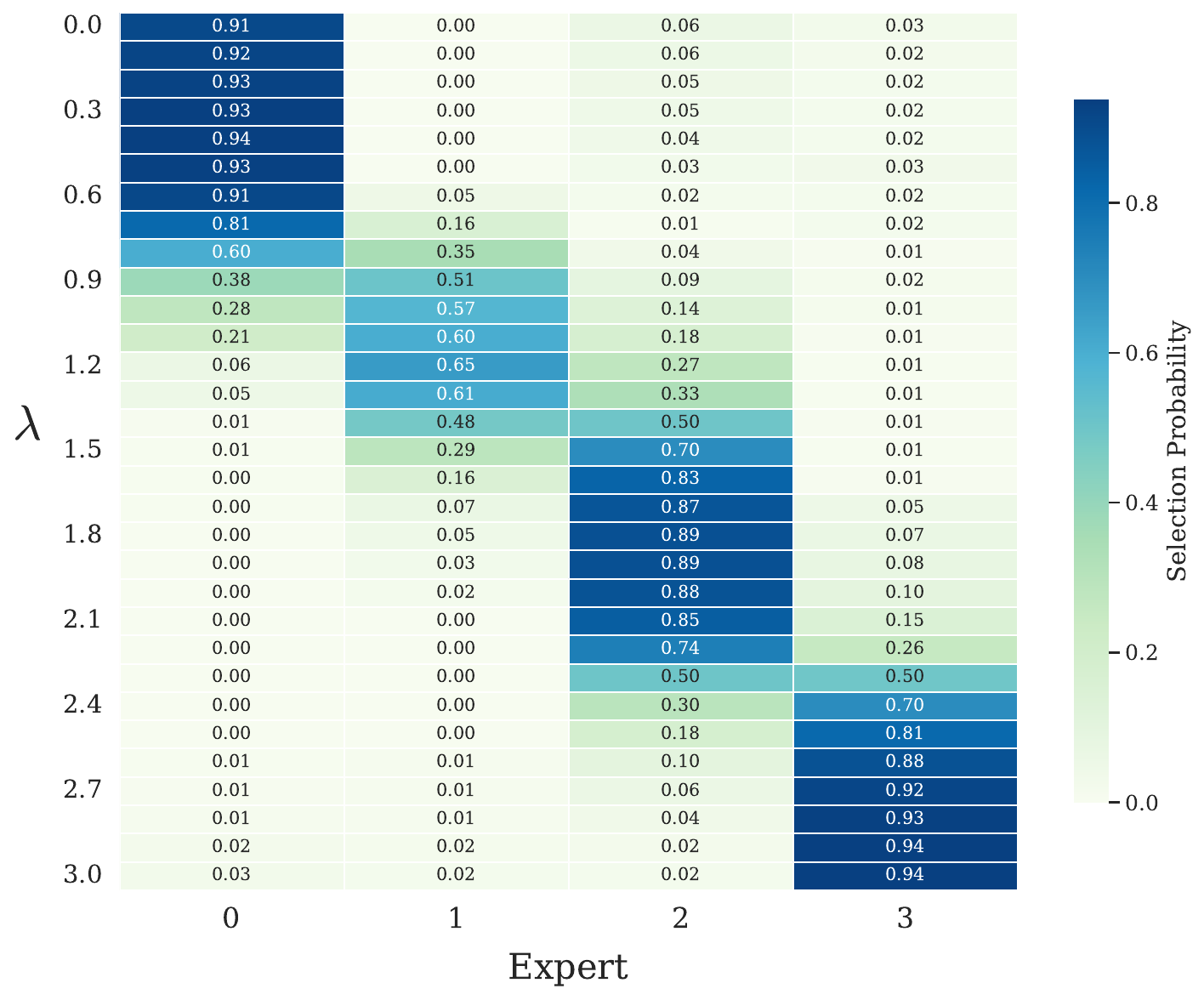}
  \caption{Expert specialization induced by the constraint signal $\lambda$ in the constraint-aware MoE on IA dataset.}
  \label{fig:moe}
\end{figure}

Fig.~\ref{fig:moe} visualizes the average expert selection probabilities as $\lambda$ varies from loose to strict constraint regimes.
When $\lambda$ is small, corresponding to a relaxed ROI constraint, the gating network predominantly activates Expert~0.
This expert is consistently assigned the highest routing probability and learns to generate aggressive incentive allocations that prioritize revenue maximization with limited regard for cost efficiency.
As $\lambda$ increases, the routing probability gradually shifts from Expert~0 toward Expert~1 and Expert~2.
This smooth transition indicates that the gating network does not treat $\lambda$ as a discrete switch, but instead continuously modulates the composition of experts to balance revenue and cost under moderate constraint pressure.
In the high-$\lambda$ regime, the routing mass is almost entirely concentrated on Expert~3, which specializes in conservative, cost-sensitive incentive generation and strongly suppresses low-efficiency actions.

Importantly, this expert specialization is not manually imposed.
All experts share the same architecture and are trained jointly, and the observed division of labor emerges naturally from conditioning the MoE on $\lambda$ during training.
These results demonstrate that the constraint-aware MoE effectively translates global constraint pressure into structured, expert-level behavioral specialization, enabling the model to adapt its incentive strategy across different ROI regimes without retraining.

\section{Synthetic ROI-Constrained Environment}
\label{app:syn}

To isolate policy controllability under long-term ROI constraints away from live-data confounders, we introduce a minimal, reproducible synthetic testbed capturing the tension between short-term incentive gains and long-term fatigue-induced ROI degradation. Each session evolves over $T$ timesteps. At step $t$, the agent observes state $s_t = (f_t, r_{t-1})$, where $f_t$ represents user fatigue and $r_{t-1} \in \{0,1\}$ is the prior engagement outcome. The agent selects a discrete incentive $a_t \in \{0,1,\dots,K\}$ with a linear cost $C_t = a_t$. 

Our fatigue model is intentionally simplified as a controlled testbed for mechanistic verification. Fatigue evolves deterministically as $f_{t+1} = \rho f_t + \eta a_t$, where $\rho \in (0,1)$ governs natural decay and $\eta > 0$ tracks accumulation. We adopt this exponential decay to align with empirical observations in advertising and recommendation: fatigue from repeated exposures dissipates rapidly at first, then tapers off gradually. 

User engagement $r_t \sim \mathrm{Bernoulli}(p_t)$ yields the instantaneous reward $R_t = r_t$. To incorporate saturation effects, fatigue scales user engagement non-linearly via a sigmoid function, yielding the success probability $p_t = \sigma(\alpha a_t - \beta f_t)$, where $\alpha > 0$ dictates the immediate utility of incentives and $\beta > 0$ governs fatigue sensitivity. Trajectory-level ROI is defined as $\mathrm{ROI} = \sum r_t / (\sum a_t + \epsilon)$. 

To build the offline dataset, we sample $10,000$ trajectories via a mixed behavior policy (greedy, random, and fixed medium incentives) to ensure broad action coverage and fair evaluation. Default parameters are set to $K=10$, $T=100$, $\alpha=0.8$, $\beta=1.2$, $\rho=0.9$, and $\eta=0.5$.

\section{Dataset Details}
\label{app:data}
\begin{itemize}[leftmargin=*, topsep=0pt, partopsep=0pt, parsep=0pt]

\item \textbf{DT}~\cite{chen2021decision}: A generative approach that casts offline RL as a conditional sequence modeling task, generating actions conditioned on the target return-to-go.
\item \textbf{CDT}~\cite{liu2023constrained}: A safe RL variant of DT that conditions generation on cost thresholds.

\item \textbf{IQL}~\cite{kostrikov2021offline}: A state-of-the-art offline RL method that mitigates out-of-distribution overestimation via expectile-based value learning.

\item \textbf{CAL}~\cite{wu2024off}: It solves the constrained problem by iteratively updating the policy and the dual variable.

\item \textbf{TREBI}~\cite{lin2023safe}: It tackles constrained policy optimization by optimizing over trajectory distributions while enforcing strict constraint satisfaction.

% \item \textbf{CAPS}~\cite{chemingui2025constraint}: It learns multiple policies under different reward–cost trade-offs and selects actions that maximize future returns subject to cost constraints.
\end{itemize}

\end{document}